\documentclass[10pt,twocolumn,letterpaper]{article}

\usepackage{cvpr}              

\usepackage{graphicx}
\usepackage{amsmath}
\usepackage{amssymb}
\usepackage{booktabs}

\usepackage{amsthm}
\usepackage{colortbl}
\usepackage{color}
\usepackage{array}
\usepackage{verbatim}
\usepackage{bm}
\usepackage{floatrow}
\usepackage{enumerate}

\usepackage{times}
\usepackage{epsfig}

\usepackage{multirow}

\usepackage{soul}
\setulcolor{red}
\sethlcolor{green}
\setstcolor{blue}

\usepackage[accsupp]{axessibility}  

\usepackage[pagebackref,breaklinks,colorlinks]{hyperref}
\usepackage{url}            

\usepackage[capitalize]{cleveref}
\crefname{section}{Sec.}{Secs.}
\Crefname{section}{Section}{Sections}
\Crefname{table}{Table}{Tables}
\crefname{table}{Tab.}{Tabs.}

\usepackage{appendix}
\usepackage{lipsum}

\def\cvprPaperID{11044} 
\def\confName{CVPR}
\def\confYear{2023}

\begin{document}

\title{A New Benchmark: On the Utility of Synthetic Data with Blender for \\ Bare Supervised Learning and Downstream Domain Adaptation}

\author{Hui Tang$^{1,2}$ and Kui Jia$^{1,}$\thanks{Corresponding author.}\\
	$^1$ South China University of Technology \quad $^2$ DexForce Co. Ltd. \\
	{\tt\small eehuitang@mail.scut.edu.cn}, {\tt\small kuijia@scut.edu.cn}
}
\maketitle

\begin{abstract}
   Deep learning in computer vision has achieved great success with the price of large-scale labeled training data. However, exhaustive data annotation is impracticable for each task of all domains of interest, due to high labor costs and unguaranteed labeling accuracy. Besides, the uncontrollable data collection process produces non-IID training and test data, where undesired duplication may exist. All these nuisances may hinder the verification of typical theories and exposure to new findings. To circumvent them, an alternative is to generate synthetic data via 3D rendering with domain randomization. We in this work push forward along this line by doing profound and extensive research on bare supervised learning and downstream domain adaptation. 
   Specifically, under the well-controlled, IID data setting enabled by 3D rendering, we systematically verify the typical, important learning insights, e.g., shortcut learning, 
   and discover the new laws of various data regimes and network architectures in generalization. 
   We further investigate the effect of image formation factors on generalization, e.g., object scale, material texture, illumination, camera viewpoint, and background in a 3D scene. 
   Moreover, we use the simulation-to-reality adaptation as a downstream task for comparing the transferability between synthetic and real data when used for pre-training, which demonstrates that synthetic data pre-training is also promising to improve real test results. 
   Lastly, to promote future research, we develop a new large-scale synthetic-to-real benchmark for image classification, termed S2RDA, which provides more significant challenges for transfer from simulation to reality. 
\end{abstract}

\section{Introduction}
\label{sec:intro}

Recently, we have witnessed considerable advances in various computer vision applications \cite{ilsvrc,fcn_ss,r_cnn}. 
However, such a success is vulnerable and expensive in that it has been limited to supervised learning methods with abundant labeled data. 
Some publicly available datasets exist certainly, which include a great mass of real-world images and acquire their labels via crowdsourcing generally. For example, ImageNet-1K \cite{imagenet} is of $1.28$M images
; MetaShift \cite{MetaShift} has $2.56$M natural images. 
Nevertheless, data collection and annotation for all tasks of domains of interest are impractical since many of them require exhaustive manual efforts and valuable domain expertise, e.g., self-driving and medical diagnosis. What's worse, the label given by humans has no guarantee to be correct, resulting in unpredictable label noise. Besides, the poor-controlled data collection process produces a lot of nuisances, e.g., training and test data aren't independent identically distributed (IID) and even have duplicate images. All of these shortcomings could prevent the validation of typical insights and exposure to new findings. 

To remedy them, one can resort to synthetic data generation via 3D rendering \cite{blenderproc}, where an arbitrary number of images can be produced with diverse values of imaging factors randomly chosen in a reasonable range, i.e., domain randomization \cite{domain_randomization}; such a dataset creation pipeline is thus very lucrative, where data with labels come for free. For image classification, Peng et al. \cite{visda2017} propose the first large-scale synthetic-to-real benchmark for visual domain adaptation \cite{survey_tl_early}, VisDA-2017; it includes $152$K synthetic images and $55$K natural ones. 
Ros et al. \cite{synthia} produce $9$K synthetic cityscape images for cross-domain semantic segmentation. Hinterstoisser et al. \cite{retail_dataset} densely render a set of $64$ retail objects for retail detection. All these datasets are customized for specific tasks in cross-domain transfer. In this work, we push forward along this line extensively and profoundly. 

The deep models tend to find simple, unintended solutions and learn shortcut features less related to the semantics of particular object classes, due to systematic biases, as revealed in \cite{shortcut_learning}. For example, a model basing its prediction on context would misclassify an airplane floating on water as a boat. The seminal work \cite{shortcut_learning} emphasizes that shortcut opportunities are present in most data and rarely disappear by simply adding more data. Modifying the training data to block specific shortcuts may be a promising solution, e.g., making image variation factors consistently distributed across all categories. To empirically verify the insight, we propose to compare the traditional fixed-dataset periodic training strategy with a new strategy of training with unduplicated examples generated by 3D rendering, under the well-controlled, IID data setting. We run experiments on three representative network architectures of ResNet \cite{resnet}, ViT \cite{ViT}, and MLP-Mixer \cite{MLP-Mixer}, which consistently show obvious advantages of the data-unrepeatable training (cf. Sec. \ref{sec:4.1})
. This also naturally validates the typical arguments of probably approximately correct (PAC) generalization 
\cite{understandingML} 
and variance-bias trade-off 
\cite{useful_things_ml}. Thanks to the ideal IID data condition enabled by the well-controlled 3D rendering, we can also discover more reliable laws of various data regimes and network architectures in generalization
. Some interesting observations are as follows. 
\begin{itemize}
	\item \textbf{Do not learn shortcuts!} 
	The test results on synthetic data without background are good enough to show that the synthetically trained models do not learn shortcut solutions relying on context clues \cite{shortcut_learning}. 
	\item \textbf{A zero-sum game.} For the data-unrepeatable training, IID and OOD (Out-of-Distribution \cite{survey_dg,M-ADA}) generalizations are some type of zero-sum game w.r.t. the strength of data augmentation. 
	\item \textbf{Data augmentations do not help ViT much!} In IID tests, ViT performs surprisingly poorly whatever the data augmentation is and even the triple number of training epochs does not improve much. 
	\item \textbf{There is always a bottleneck from synthetic data to OOD/real data}. Here, increasing data size and model capacity brings no more benefits, and domain adaptation \cite{survey_da} to bridge the distribution gap is indispensable except for evolving the image generation pipeline to synthesize more realistic images.
\end{itemize}

Furthermore, we comprehensively assess image variation factors, e.g., object scale, material texture, illumination, camera viewpoint, and background in a 3D scene. We then find that to generalize well, deep neural networks must \emph{learn to ignore non-semantic variability}, which may appear in the test. To this end, sufficient images with different values of one imaging factor should be generated to learn a robust, unbiased model, proving the necessity of sample diversity for generalization \cite{survey_data_aug,survey_dg,domain_randomization}. We also observe that \emph{different factors and even their different values have uneven importance to IID generalization}, implying that the under-explored weighted rendering \cite{AutoSimulate} is worth studying.

Bare supervised learning on synthetic data results in poor performance in OOD/real tests, and pre-training and then domain adaptation can improve. Domain adaptation (DA) \cite{survey_da} is a hot research area, which aims to make predictions for unlabeled instances in the target domain by transferring knowledge from the labeled source domain. To our knowledge, there is little research on pre-training for DA \cite{study_pretrain_for_DA} (with real data). We thus use the popular simulation-to-real classification adaptation \cite{visda2017} as a downstream task, study the transferability of synthetic data pre-trained models by comparing with those pre-trained on real data like ImageNet and MetaShift
. We report results for several representative DA methods \cite{dann,mcd,rca,srdc,disclusterda} on the commonly used backbone, and our experiments yield some surprising findings. 
\begin{itemize}
	\item \textbf{The importance of pre-training for DA.} DA fails without pre-training (cf. Sec. \ref{sec:4.3.2}). 
	\item \textbf{Effects of different pre-training schemes.} Different DA methods exhibit different relative advantages under different pre-training data. The reliability of existing DA method evaluation criteria is unguaranteed. 
	\item \textbf{Synthetic data pre-training vs. real data pre-training.} Synthetic data pre-training is better than pre-training on real data in our study. 
	\item \textbf{Implications for pre-training data setting.} Big Synthesis Small Real is worth researching. Pre-train with target classes first under limited computing resources. 
	\item \textbf{The improved generalization of DA models.} Real data pre-training with extra non-target classes, fine-grained target subclasses, or our synthesized data added for target classes helps DA. 
\end{itemize}

Last but not least, we introduce a new, large-scale synthetic-to-real benchmark for classification adaptation (S2RDA), which has two challenging tasks S2RDA-49 and S2RDA-MS-39. 
S2RDA contains more categories, more realistically synthesized source domain data coming for free, and more complicated target domain data collected from diverse real-world sources, setting a more practical and challenging benchmark for future DA research. 

\section{Related Works}
\label{sec:related_works}

\noindent\textbf{Real Datasets.} A lot of large-scale real datasets \cite{imagenet,imagenet-21k,JFT-300M,MetaShift,StylizedImageNet,ms-coco,L-Bird,WebFG} have harnessed and organized the explosive image data from the Internet or real world for deep learning of meaningful visual representations. For example, ImageNet \cite{imagenet} is a large-scale image database 
consisting of $1.28$M images from $1$K common object categories, and serves as the primary dataset for pre-training deep models for vision tasks. Barbu et al. \cite{ObjectNet} collect a large real-world test set for more realistic object recognition, ObjectNet, which has $50$K images and is bias-controlled. 
MetaShift \cite{MetaShift} of $2.56$M natural images ($\sim400$ classes) is formed by context guided clustering of the images from GQA \cite{CleanedVG}
. 


\noindent\textbf{Synthetic Datasets.} Thanks to 3D rendering \cite{blenderproc} and domain randomization \cite{domain_randomization}, synthetic data with increased sample diversity can be generated for free now, facilitating various vision tasks. 
VisDA-2017 \cite{visda2017} is a large-scale benchmark dataset for cross-domain object classification, focusing on the simulation-to-reality shift, with $152$K synthetic images and $55$K natural ones across $12$ categories. For cross-domain semantic segmentation, Ros et al. \cite{synthia} produce $9$K synthetic images by rendering a virtual city using the Unity engine; 
many other works \cite{adv_tuned_scene_gen,model_driven_sim,sim4val,hypersim} focus on computer graphics simulations to synthesize outdoor or indoor scene images. 
For retail detection, Hinterstoisser et al. \cite{retail_dataset} densely render $64$ retail objects and a large dataset of 3D background models
. For car detection, domain randomization is also utilized and developed in \cite{DR4ObjectDetection,SDR4CarDetection}
. 

\noindent\textbf{Domain Adaptation.} Domain adaptation is a developing field with a huge diversity of approaches. A popular strategy is to explicitly model and minimize the distribution shift between the source and target domains \cite{dann,mcd,rca,SymNets,tpn,vicatda}, such that the domain-invariant features can be learned
. 
Differently, works of another emerging strategy \cite{srdc,disclusterda,tat,bnm} take steps towards implicit domain adaptation, without explicit feature alignment
. 
In this work, we consider these representative DA methods for the empirical study, and broader introductions to the rich literature are provided in \cite{survey_da,U-WILDS}.



\section{Data Synthesis via Domain Randomization}
\label{sec:renderer}

We adopt the widely used 3D rendering \cite{blenderproc} in a simulator for data synthesis and generate synthetic RGB images for model training. To increase sample diversity for better generalization, we apply domain randomization \cite{domain_randomization} during rendering, whose efficacy has been demonstrated in various applications \cite{visda2017,DR4ObjectDetection,SDR4CarDetection}. 
Specifically, we start by sampling a 3D object model from one specific class of interest from ShapeNet repository \cite{shapenet2015} and place it in a blank scene; next, we set the lighting condition with a point source of randomized parameters and place the camera at random positions on the surface of a sphere of random radius, which has lens looking at the object and the intrinsic resolution of 256$\times$256; next, we apply random materials and textures to the object; then, we use an RGB renderer to take pictures from different camera viewpoints in the configured scene; finally, the rendered images are composed over a background image chosen at random from open resources. The synthesized images with the automatically generated ground truth class labels are used as low-cost training data. 
%

\begin{figure}[!t]
	\centering
	\includegraphics[width=1.0\textwidth]{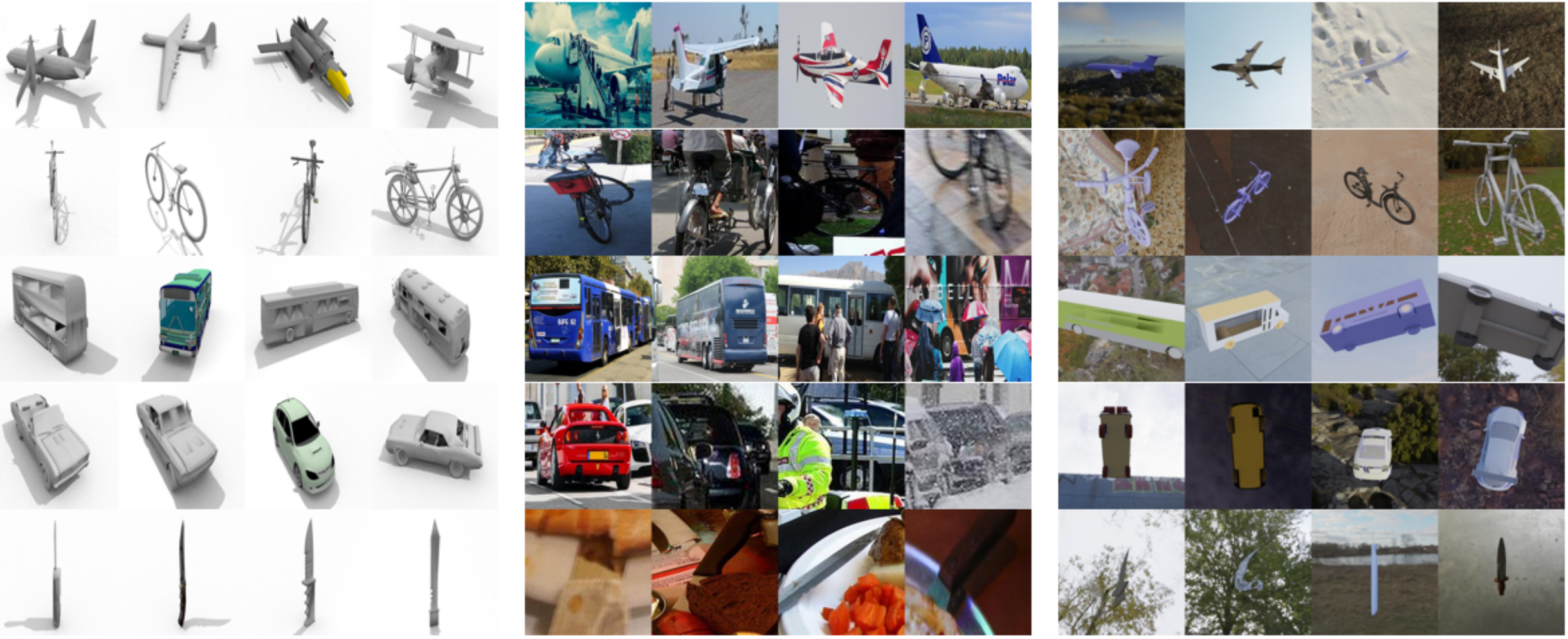}
	\vspace{-0.5cm}
	\caption{Sample images from the training (left) and validation (middle) domains of VisDA-2017 and our synthesized data (right).}
	\label{fig:show_imgs_visda_vs_ours}
\end{figure}

The changing ranges of image variation factors in the 3D scene are as follows. 
\textbf{(1)} The scale of the object in an image depends on its distance to the camera, namely the radius of a sphere on whose surface the camera is located. The radius is ranged in $[0.8,2.4]$. 
\textbf{(2)} The material texture of an object is from CCTextures \cite{CCTextures}
, which contains actual images. 
\textbf{(3)} The point light is on a shell centered at $[1, 2, 3]$, whose radius and elevation are ranged in $[1,7]$ and $[15, 70]$ respectively. 
\textbf{(4)} The camera lies on the surface of a sphere centered at $[0,0,0]$ (also the object center) and its azimuth and elevation are from $0^\circ$ to $360^\circ$. 
\textbf{(5)} The background images are from Haven \cite{Haven}
, which includes environment HDRs.

\noindent\textbf{Remarks.} It is noteworthy that VisDA-2017 \cite{visda2017} generates synthetic images by rendering 3D models just under varied camera angles and lighting conditions. Differently, we vary the values of much more image variation factors, leading to more realistic and diverse samples, as illustrated in Fig. \ref{fig:show_imgs_visda_vs_ours}.

\section{Experiment and Evaluation}
\label{sec:exp&eval}

We empirically demonstrate the utility of our synthetic data for supervised learning and downstream transferring by exploring the answers to the following questions:

{
	\noindent{$\bullet$ \enspace} Can we utilize synthetic data to verify typical theories and expose new findings? What will we find when investigating the learning characteristics and properties of our synthesized new dataset comprehensively?
	
	\noindent{$\bullet$ \enspace} Can a model trained on non-repetitive samples converge? If it could, how will the new training strategy perform when compared to fixed-dataset periodic training? Can the comparison provide any significant intuitions for shortcut learning and other insights?
	
	\noindent{$\bullet$ \enspace} How will the image variation factors in domain randomization affect the model generalization? What new insights can the study provide for 3D rendering?
	
	\noindent{$\bullet$ \enspace} Can synthetic data pre-training be on par with real data pre-training when applied to downstream synthetic-to-real classification adaptation? How about large-scale synthetic pre-training with a small amount of real data?
	
	\noindent{$\bullet$ \enspace} Is our S2RDA benchmark more challenging and realistic? How does it differ from VisDA-2017?
}

\subsection{Empirical Study on Supervised Learning}
\label{sec:4.1}


\noindent\textbf{Data Settings.} We use the $10$ object classes common in ShapeNet and VisDA-2017 for the empirical study
. We term the synthetic and real domains of the $10$ classes in VisDA-2017 as SubVisDA-10
. For the traditional fixed-dataset periodic training, we generate $12$K synthetic images in each class and train the model on the dataset with fixed size epoch by epoch. For our used sample-unrepeatable training, we have mutually exclusive batches of synthesized samples per iteration. At inference, we evaluate the learned model on three types of test data: IID data of $6$K samples per class which follow the same distribution as the synthesized training data, IID data of $60$K images without background to examine the dependency of network predictions on contexts, and OOD data, i.e., real images from SubVisDA-10. For training, we consider three data augmentation strategies with different strengths: no augmentation which has only the center crop operation, weak augmentation based on pixel positions such as random crop and flip \cite{imagenet}, and strong augmentation which transforms both position and value of pixels in an image, e.g., random resized crop and color jitter
\cite{SimCLR}. In the test phase, we use no augmentation. 


\noindent\textbf{Implemental Details.} We adopt the standard cross-entropy loss function for pattern learning. We do experiments using ResNet-50 \cite{resnet}, ViT-B \cite{ViT}, and Mixer-B \cite{MLP-Mixer} as the backbone
. We train the model from scratch for $200$K iterations and use the SGD optimizer with batch size $64$ and the cosine learning rate schedule to update network parameters. 
We report the overall accuracy (Acc. \%) or mean class precision (Mean \%) at the same fixed random seed across all experiments. Other settings are detailed in the appendix.


\subsubsection{Results}

\noindent\textbf{Fixed-Dataset Periodic Training vs. Training on Non-Repetitive Samples.} 
Under the ideal IID data condition enabled by 3D rendering, we empirically verify significant insights on shortcut learning \cite{shortcut_learning}, PAC generalization \cite{understandingML}, and variance-bias trade-off \cite{useful_things_ml}, by making comparisons between fixed-dataset periodic training and training on non-repetitive samples. Results are shown in Table \ref{tab:fd_vs_ut} and Figs. \ref{fig:learning_curves} and A1 (learning process in the appendix). We highlight several observations below. 
\textbf{(1)} With more training data of increased sample diversity, the data-unrepeatable training exhibits \emph{higher generalization accuracy and better convergence performance} than the fixed-dataset training. 
\textbf{(2)} To intuitively understand what the models have learned, we visualize the saliency/attention maps in Figs. A2-A5. We observe that all models attend to image regions from global (context) to local (object) as the learning process proceeds; the data-unrepeatable training achieves \emph{qualitative improvements} over the fixed-dataset training. 
\textbf{(3)} Our synthesized data used for training yield \emph{higher OOD test accuracy} than SubVisDA-10 as they share more similarities to the real data, as shown in Fig. \ref{fig:show_imgs_visda_vs_ours}.
\textbf{(4)} The fixed-dataset training displays \emph{overfitting phenomenons} whilst the data-unrepeatable training \emph{does not} (cf. (a-d) in Fig. A1), since the former samples training instances from an empirical distribution with high bias and low variance, and thus cannot perfectly generalize to the unseen test instances sampled from the true distribution. 
\textbf{(5)} With strong data augmentation, the data-unrepeatable training has the test results on IID w/o BG data not only at their best but also better than those on IID data, implying that the trained models \emph{do not learn shortcut solutions that rely on context clues in the background}.

\begin{table}[t]
	\caption{Training on a fixed dataset vs. non-repetitive samples. FD: Fixed Dataset, True (T) or False (F). DA: Data Augmentation, None (N), Weak (W), or Strong (S). BG: BackGround.
	}
	\label{tab:fd_vs_ut}
	\centering
	\resizebox{0.8\linewidth}{!}{
		\begin{tabular}{lcccccc}
			\toprule
			\multirow{2}{*}{Data} & \multirow{2}{*}{FD} & \multirow{2}{*}{DA} & IID & IID w/o BG & \multicolumn{2}{c}{OOD} \\
			&    &    & Acc./Mean & Acc./Mean & Acc. & Mean \\
			\midrule
			\multicolumn{7}{c}{\textbf{Backbone: ResNet-50 (23.53M)}} \\
			SubVisDA-10 & T & N & 11.25 & 11.72 & 22.02 & 14.71 \\
			Ours & T & N & 87.63 & 78.55 & 23.35 & 23.36 \\
			Ours & F & N & \textbf{98.19} & \textbf{96.39} & \textbf{25.04} & \textbf{26.05} \\
			\cmidrule(r){2-5}
			SubVisDA-10 & T & W & 12.31 & 13.53 & 25.95 & 16.83 \\
			Ours & T & W & 95.54 & 91.37 & 23.97 & 22.89 \\
			Ours & F & W & \textbf{98.10} & \textbf{96.35} & \textbf{27.47} & \textbf{27.49} \\
			\cmidrule(r){2-5}
			SubVisDA-10 & T & S & 17.39 & 20.32 & 33.07 & 27.48 \\
			Ours & T & S & 94.86 & 95.33 & 42.24 & 41.73 \\
			Ours & F & S & \textbf{96.26} & \textbf{96.50} & \textbf{42.82} & \textbf{42.25} \\		
			\midrule
			\multicolumn{7}{c}{\textbf{Backbone: ViT-B (85.78M)} \quad \dag: Training for $600$K iterations} \\
			SubVisDA-10 & T & N & 12.68 & 11.30 & 24.28 & 17.81 \\
			Ours & T & N & 68.51 & 61.50 & 26.65 & 24.13 \\
			Ours\dag & T & N & 70.58 & 62.15 & 26.57 & 24.23 \\
			Ours & F & N & \textbf{76.34} & \textbf{71.46} & \textbf{30.10} & \textbf{26.93} \\
			\cmidrule(r){2-5}
			SubVisDA-10 & T & W & 11.77 & 11.20 & 26.53 & 19.22 \\
			Ours & T & W & 72.79 & 67.46 & \textbf{30.04} & 26.45 \\
			Ours & F & W & \textbf{73.93} & \textbf{68.59} & 29.92 & \textbf{26.80} \\
			\cmidrule(r){2-5}
			SubVisDA-10 & T & S & 14.45 & 12.89 & 31.52 & 23.74 \\
			Ours & T & S & 62.85 & 63.96 & \textbf{31.79} & \textbf{26.56} \\
			Ours & F & S & \textbf{64.26} & \textbf{64.30} & 30.89 & 26.28 \\	
			\midrule
			\multicolumn{7}{c}{\textbf{Backbone: Mixer-B (59.12M)}} \\
			SubVisDA-10 & T & N & 12.85 & 15.17 & 21.56 & 17.02 \\
			Ours & T & N & 66.05 & 57.66 & 21.85 & 21.22 \\
			Ours & F & N & \textbf{90.22} & \textbf{85.86} & \textbf{28.54} & \textbf{27.98} \\
			\cmidrule(r){2-5}
			SubVisDA-10 & T & W & 13.99 & 23.12 & 27.67 & 19.86 \\
			Ours & T & W & 78.43 & 71.48 & 27.15 & 26.01 \\
			Ours & F & W & \textbf{90.32} & \textbf{86.13} & \textbf{29.11} & \textbf{29.49} \\
			\cmidrule(r){2-5}
			SubVisDA-10 & T & S & 14.88 & 24.85 & 33.19 & 26.12 \\
			Ours & T & S & 81.72 & 83.06 & \textbf{36.57} & 33.43 \\
			Ours & F & S & \textbf{84.16} & \textbf{85.25} & 36.50 & \textbf{33.75} \\	
			\bottomrule
		\end{tabular}
	}\vspace{-0.3cm}
\end{table}

\begin{figure*}[t]
	\centering
	\subfloat[Test loss (IID)]{
		\begin{minipage}[t]{0.24\textwidth}
			\includegraphics[height=1.05in]{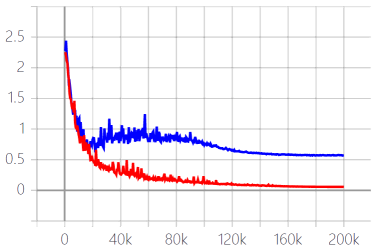}
			\label{fig:learning_curves:2}
		\end{minipage}
	}
	\subfloat[Test loss (IID)]{
		\begin{minipage}[t]{0.24\textwidth}
			\includegraphics[height=1.05in]{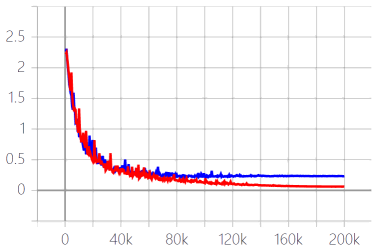}
			\label{fig:learning_curves:10}
		\end{minipage}
	}
	\subfloat[Test loss (IID)]{
		\begin{minipage}[t]{0.24\textwidth}
			\includegraphics[height=1.05in]{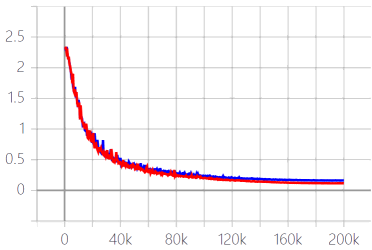}
			\label{fig:learning_curves:18}
		\end{minipage}
	}
	\subfloat[Test acc. (OOD)]{
		\begin{minipage}[t]{0.28\textwidth}
			\includegraphics[height=1.05in]{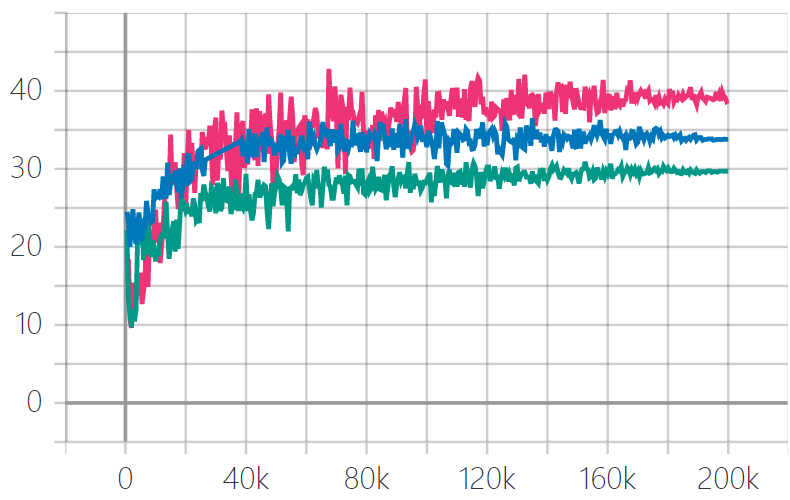}			
		\end{minipage}
	\label{fig:comparison_of_3models}
	}
	\vspace{-0.3cm}
	\caption{Learning process. \textbf{(a-c):}  Training ResNet-50 on a fixed dataset (\textcolor{blue}{blue}) or non-repetitive samples (\textcolor{red}{red}) for no, weak, and strong data augmentations. \textbf{(d):} Training ResNet-50 (\textcolor{red}{red}), ViT-B (\textcolor{green}{green}), and Mixer-B (\textcolor{blue}{blue}) on non-repetitive samples with strong data augmentation.}
	\label{fig:learning_curves}
\end{figure*}

\noindent\textbf{Evaluating Various Network Architectures.} 
In addition to Table \ref{tab:fd_vs_ut}, we also show the learning process of various network architectures in Figs. \ref{fig:comparison_of_3models} and A6. We take the following interesting observations. 
\textbf{(1)} On the fixed-dataset training and IID tests, 
ViT-B \emph{performs surprisingly poorly} whatever the data augmentation is, when compared with ResNet-50; even the triple number of training epochs does not work as well as expected (e.g., in \cite{ViT}). 
%
\textbf{(2)} When training on non-repetitive images without strong data augmentation, ViT-B and Mixer-B perform better than ResNet-50 in OOD tests whereas they perform much worse with strong data augmentation. Maybe they are more suitable for handling data with a certain (or smaller) range of diversity. Namely, \emph{different network architectures have different advantages for different data augmentations}, suggesting that neural architecture search (NAS) should also consider the search for data augmentation.
\textbf{(3)} With strong data augmentation, ResNet-50 fits best and shows the best convergence, though it has a more volatile learning process for the OOD test (cf. Figs. \ref{fig:comparison_of_3models} and A6).
\textbf{(4)} 
ResNet-50 produces more accurate saliency map visualizations, where the attended regions are semantically related (cf. Figs. A2-A5).

\begin{figure}[t]
	\centering
	\begin{subfigure}{0.45\textwidth}
		\flushleft
		\includegraphics[height=1.01in]{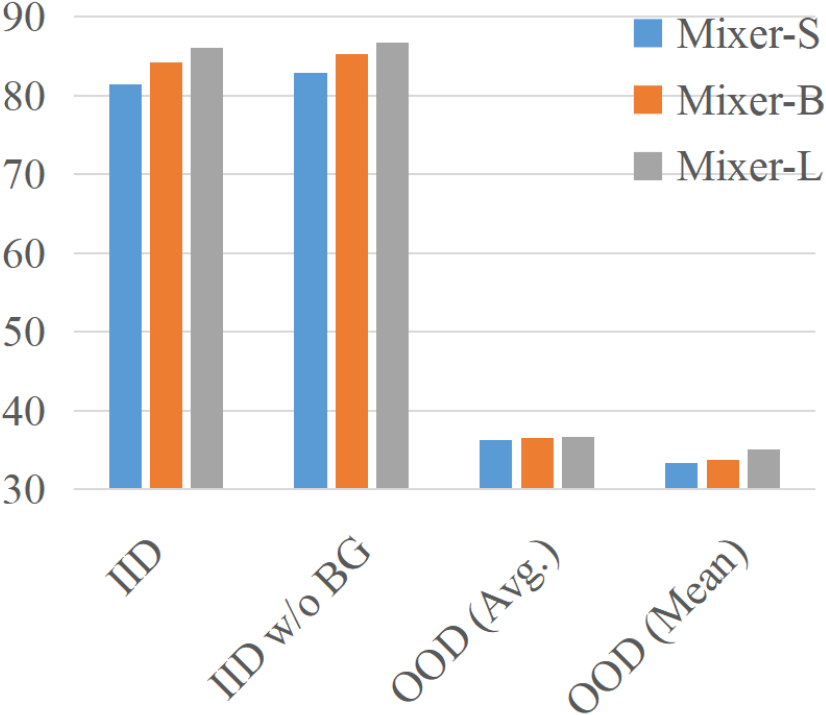}
		\label{fig:accs:diff_network_archs}
\end{subfigure}
\hspace{-3mm}
\begin{subfigure}{0.55\textwidth}
	\flushleft
	\includegraphics[height=1.01in]{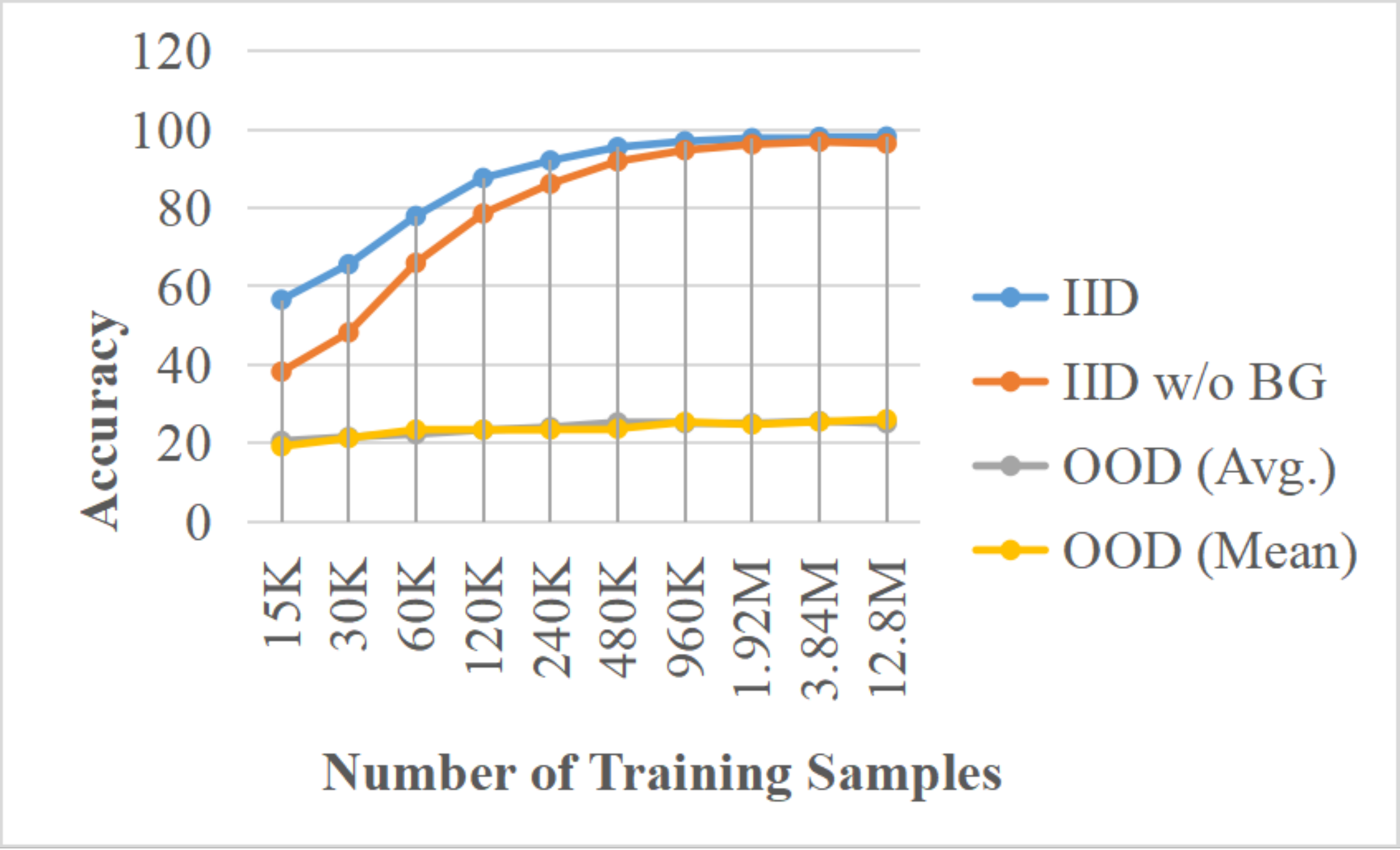}
	\label{fig:accs:num_of_training_samples}
\end{subfigure}
\vspace{-0.3cm}
\caption{Generalization accuracy w.r.t. model capacity (\textbf{a/left}) or training data quantity (\textbf{b/right}). }
\label{fig:accs}
\end{figure}

\noindent\textbf{Impact of Model Capacity.} When Mixer is used as the backbone for data-unrepeatable training with strong data augmentation, we do experiments by varying the number of layers in $[8, 12, 24]$, i.e., Mixer-S ($18.02$M), Mixer-B ($59.12$M), and Mixer-L ($207.18$M). The results are shown in Fig. \ref{fig:accs}\textcolor{red}{a}. Although better results are achieved by higher capacity models, the performance gain is \emph{less and less significant}
. It also suggests that given a specific task, one model always has a bottleneck, which may be broken by adjusting the training data and learning algorithm.

\noindent\textbf{Impact of Training Data Quantity.} 
We do experiments by doubling the number of training samples and the last 12.8M is the upper bound
. We use the fixed-dataset periodic training with ResNet-50 and no data augmentation, and show the results in Fig. \ref{fig:accs}\textcolor{red}{b}. 
\textbf{(1)} In IID tests, as the number of training samples is continuously doubled, the performance gets better and is near perfect finally. 
\textbf{(2)} In real OOD tests, the performance gain is slighter and along the last 4 numbers, the performance gain almost disappears under our data regime. 
It demonstrates that simply generating more synthesized images may \emph{get stuck} at last and one can resort to domain adaptation approaches to reduce the distribution shift or more realistic simulation for image synthesis.

\noindent\textbf{Impact of Data Augmentations.} 
Data augmentation plays an important role in deep learning and we separately analyze its impact. There are a few noteworthy findings in Table \ref{tab:fd_vs_ut}. 
\textbf{(1)} For training ResNet-50 on a fixed dataset, weak augmentation can enable the learnability from our synthesized data to IID data (e.g., $>95\%$); from synthetic to IID w/o BG, strong augmentation is necessary
; from synthetic to OOD, strong augmentation has the highest learnability. These observations enlighten us: \emph{given a limited set of training data, is there necessarily some kind of data augmentation that makes it learnable from training to test?}
\textbf{(2)} For the data-unrepeatable training, the results in IID tests \emph{get worse} while those in OOD tests \emph{get better} when strengthening the data augmentation, since the distribution of strongly augmented training data \emph{differs} from that of IID test data but is \emph{more similar} to that of OOD test data. 

\begin{table}
\caption{Fix vs. randomize image variation factors (ResNet-50).}
\label{tab:assess_img_var_factors}
\centering
\resizebox{0.91\textwidth}{!}{
	\begin{tabular}{lcclcc}
		\toprule
		\multicolumn{3}{c}{\textbf{Object Scale}} & \multicolumn{3}{c}{\textbf{Material Texture}} \\
		Value & IID & IID w/o BG & Value & IID & IID w/o BG \\
		\cmidrule(r){1-3}\cmidrule(r){4-6}
		
		1   & 68.77 & 58.00 & Metal & 79.58 & 68.78 \\
		1.5 & 80.80 & 72.22 & Plastic & 50.29 & 46.82 \\
		2   & 77.61 & 70.10 & Fingerprints & 50.35 & 62.27 \\
		Mix & 87.12 & 77.55 & Moss & 68.62 & 63.93 \\
		\midrule
		\multicolumn{3}{c}{\textbf{Illumination}} & \multicolumn{3}{c}{\textbf{Camera Viewpoint}} \\
		Value & IID & IID w/o BG & Value & IID & IID w/o BG \\
		\cmidrule(r){1-3}\cmidrule(r){4-6}
		Location 1 & 86.48 & 76.02 & Location 1 & 24.60 & 26.56 \\
		Location 2 & 86.60 & 76.75 & Location 2 & 27.21 & 28.88 \\
		Radius & 86.91 & 78.83 & Location 3 & 32.82 & 32.76 \\
		Elevation & 87.12 & 77.39 & Location 4 & 33.79 & 33.07 \\
		\midrule
		\multicolumn{3}{c}{\textbf{Background}} & \multicolumn{3}{c}{\textbf{Full Randomization}} \\
		Value & IID & IID w/o BG & Value & IID & IID w/o BG \\
		\cmidrule(r){1-3}\cmidrule(r){4-6}
		No Background & 17.68 & \textbf{94.75} & Random & \textbf{87.63} & 78.55 \\
		\bottomrule
	\end{tabular}
}\vspace{-0.3cm}
\end{table}

\begin{table*}[t]
\caption{Domain adaptation performance on SubVisDA-10 with varied pre-training schemes (ResNet-50). 
	$\bigstar$: Official checkpoint. Green or red: Best Acc. or Mean in each row (among compared DA methods). Ours w. SelfSup: Supervised pre-training + contrastive learning \cite{SimCLR}.}
\label{tab:compare_pretrain_data}
\centering
\resizebox{0.76\linewidth}{!}{
	\begin{tabular}{lcc >{\columncolor[gray]{0.78}}c c >{\columncolor[gray]{0.78}}c c >{\columncolor[gray]{0.78}}c c >{\columncolor[gray]{0.78}}c c >{\columncolor[gray]{0.78}}c c >{\columncolor[gray]{0.78}}c c}
		\toprule
		\multirow{2}{*}{Pre-training Data} & \multirow{2}{*}{\# Iters} & \multirow{2}{*}{\# Epochs} & \multicolumn{2}{c}{No Adaptation} & \multicolumn{2}{c}{DANN} & \multicolumn{2}{c}{MCD} & \multicolumn{2}{c}{RCA} & \multicolumn{2}{c}{SRDC} & \multicolumn{2}{c}{DisClusterDA} \\
		& & & Acc. & Mean & Acc. & Mean & Acc. & Mean & Acc. & Mean & Acc. & Mean & Acc. & Mean \\
		\midrule
		No Pre-training & - & - & \hl{23.89} & 14.21 & 22.30 & \ul{17.72} & 17.99 & 16.20 & 19.15 & 15.19 & 19.58 & 15.92 & 20.87 & 17.31 \\
		
		Ours & 200K & 107 & 47.73 & 42.96 & 47.91 & 48.94 & \hl{55.23} & \ul{56.86} & 54.27 & 52.72 & 44.70 & 47.45 & 54.09 & 54.91 \\
		
		Ours w. SelfSup & 200K & 107 & 47.80 & 42.81 & 47.25 & 48.32 & \hl{56.71} & \ul{58.33} & 53.44 & 53.31 & 40.21 & 40.50 & 54.37 & 54.15 \\
		
		\midrule
		
		SynSL & 200K & 1 & 47.50 & 44.12 & 47.41 & 49.48 & \hl{55.06} & \ul{56.92} & 53.61 & 53.50 & 36.30 & 37.57 & 53.10 & 54.88 \\
		
		SynSL & 1.2M & 6 & 51.22 & 48.57 & 55.90 & 56.50 & \hl{64.52} & \ul{67.70} & 58.19 & 59.67 & 51.87 & 52.32 & 61.32 & 63.70 \\
		
		SynSL & 2.4M & 12 & 53.47 & \textbf{53.84} & 59.59 & 59.11 & \hl{65.55} & \ul{68.83} & 60.47 & 61.19 & 55.17 & 58.10 & 63.62 & 64.89 \\
		
		SynSL & 4.8M & 24 & 55.02 & 53.72 & 60.55 & \textbf{60.78} & \hl{\textbf{65.69}} & \ul{\textbf{69.53}} & 60.33 & 60.05 & 55.80 & 58.36 & 64.01 & \textbf{65.36} \\
		
		\midrule
		
		SubImageNet & 200K & 499 & 42.74 & 37.16 & 49.64 & 45.43 & 54.88 & 52.12 & \hl{58.24} & \ul{55.45} & 56.78 & 51.24 & 56.21 & 51.09 \\
		
		Ours+SubImageNet & 200K & 88 & 49.61 & 47.88 & 55.35 & 56.22 & 60.90 & 62.16 & 61.11 & 60.31 & 60.07 & 61.74 & \hl{62.22} & \ul{62.47} \\
		
		ImageNet-990 & 200K & 10 & 31.91 & 26.31 & 34.68 & 32.29 & 39.48 & 37.84 & \hl{45.10} & \ul{43.10} & 43.69 & 40.95 & 41.56 & 39.40 \\
		
		ImageNet-990+Ours & 200K & 9 & 36.53 & 30.58 & 38.15 & 35.22 & 42.38 & 41.84 & \hl{46.19} & \ul{43.45} & 45.87 & 42.95 & 42.07 & 39.40 \\
		
		\midrule
		
		ImageNet & 200K & 10 & 40.37 & 33.25 & 42.57 & 40.22 & 49.04 & 47.86 & \hl{52.36} & \ul{47.90} & 51.62 & 47.88 & 49.29 & 46.17 \\
		
		ImageNet & 1.2M & 60 & 54.69 & 51.27 & 58.50 & 56.02 & \hl{65.28} & \ul{65.88} & 62.69 & 60.28 & 60.33 & 55.00 & 62.28 & 61.00 \\
		
		ImageNet & 2.4M & 120 & 53.84 & 47.55 &	58.45 & 55.38 & \hl{65.27} & \ul{65.38} & 61.65 & 60.82 & 61.65 & 56.30 & 62.02 & 60.46 \\
		
		ImageNet$^\bigstar$ & 600K & 120 & \textbf{57.10} & 51.83 & \textbf{61.92} & 58.75 & 64.59 & 64.87 & \textbf{67.72} & \ul{\textbf{66.17}} & \hl{\textbf{69.00}} & \textbf{64.92} & \textbf{68.35} & 64.65 \\
		
		\midrule
		
		MetaShift & 200K & 5 & 38.18 & 30.31 & 38.29 & 34.04 & 45.39 & \ul{43.63} & \hl{45.93} & 42.67 & 42.83 & 38.02 & 40.17 & 35.09 \\
		
		MetaShift & 1.2M & 30 & 48.00 & 39.99 & 53.00 & 48.17 & \hl{64.04} & \ul{61.30} & 53.97 & 51.09 & 48.69 & 44.49 & 60.28 & 57.15 \\
		
		MetaShift & 2.4M & 60 & 47.24 & 39.21 & 58.41 & 53.85 & 61.10 & 58.52 & 58.64 & 55.35 & 51.71 & 47.29 & \hl{62.71} & \ul{60.18} \\
		
		\bottomrule
	\end{tabular}
}\vspace{-0.3cm}
\end{table*}

\subsection{Assessing Image Variation Factors}
\label{sec:4.2}

To know \emph{how individual image variation factors in domain randomization affect the model generalization}, we do an ablation study by fixing one of them at a time. For each variant, we accordingly synthesize $120$K  training images and train a ResNet-50 from scratch with no data augmentation. Other implementation details are the same as those in Sec. \ref{sec:4.1}. We compare the one-fixed variants with the baseline of full randomization and report results in Table \ref{tab:assess_img_var_factors}.

\noindent\textbf{Object Scale.} The scale of target object in an image is changed by modifying the distance between the object and camera in the virtual 3D scene, whose value is set as $1$, $1.5$, $2$, or a mix of the three. We can observe that when the object scale is fixed to one value, the recognition accuracy in IID and IID w/o BG tests drops significantly, e.g., by $18.86\%$ and $20.55\%$ respectively; setting the object-to-camera distance as $1.5$ achieves the best performance among the three scales; mixing the three scales restores most of the baseline performance. The observations show that \emph{decreasing the sample diversity would damage the generalization performance; different scales have different importance.}

\noindent\textbf{Material Texture.} We fix the material texture of target object in an image as 
metal, plastic, fingerprints, or moss. We observe that compared to full randomization, fixing the material texture degenerates the performance largely in IID and IID w/o BG tests, e.g., by $37.34\%$ and $31.73\%$ respectively
. 

\noindent\textbf{Illumination.} We change the illumination by fixing the light location to [$5$, $-5$, $5$] or [$4$, $5$, $6$] or narrowing the range of radius or elevation to [$3$, $4$] and [$20$, $30$]. We find that the location of light source has a greater influence than its radius and elevation. Compared with other factors, illumination has much less impact on class discrimination. 

\noindent\textbf{Camera Viewpoint.} Recall that the camera always looks at the target object. We change the viewpoint of camera by placing it in different 3D locations: [0, 1, 1], [0, -1, 1], [1, 0, 1], or [-1, 0, 1]. We can see that fixing the camera viewpoint makes the results in IID and IID w/o BG tests deteriorate greatly, e.g., by $63.03\%$ and $51.99\%$ respectively.

\noindent\textbf{Background.} When the background is lacking in an image, the accuracy in the IID test suffers an abrupt decrease of $69.95\%$ while that in the IID w/o BG test improves by $16.2\%$ due to reduced distribution shift
. Among 5 factors, the background is \emph{the most important} for IID generalization.

\noindent\textbf{Remarks.} Different rendering variation factors and even their different values have uneven importance to model generalization. It suggests that the under-explored direction of weighted rendering \cite{AutoSimulate} is worth studying, and our results in Table \ref{tab:assess_img_var_factors} provide preliminary guidance/prior knowledge for learning the distributions of variation factors
.

\subsection{Exploring Pre-training for Domain Adaptation}
\label{sec:4.3}


\noindent\textbf{Data Settings.} The data used for pre-training can be selected in several datasets and their variants: \textbf{(1)} our synthesized $120$K images of the $10$ object classes shared by SubVisDA-10 (Ours), \textbf{(2)} our synthesized $12.8$M images of the $10$ classes (for supervised learning, termed SynSL), \textbf{(3)} the subset collecting examples of the $10$ classes from ImageNet \cite{imagenet} (25,686 images, termed SubImageNet), \textbf{(4)} our synthesized $120$K images combined with SubImageNet, 
\textbf{(5)} ImageNet-990, where the fine-grained subclasses for each of the $10$ classes are merged into one, 
\textbf{(6)} ImageNet-990 combined with our 120K synthetic images, 
\textbf{(7)} the full set of ImageNet (1K classes), and \textbf{(8)} MetaShift \cite{MetaShift} (2.56M)
. For fine-tuning, we use domain adaptation (DA) on SubVisDA-10 as the downstream task, which comprises $130,725$ labeled instances in the source domain and $46,697$ unlabeled ones in the target domain. We follow the standard DA training protocol \cite{dann,mcd}
. We report classification results of overall accuracy (Acc. \%) and mean class precision (Mean \%) on the target domain under a fixed random seed.


\noindent\textbf{Implemental Details.} 
For domain adaptation, we use DANN \cite{dann}, MCD \cite{mcd}, RCA \cite{rca}, SRDC \cite{srdc}, and DisClusterDA \cite{disclusterda} as baselines. 
We closely follow the specific algorithms in the respective papers of these baselines. 
We use a pre-trained ResNet-50 as the base model
. 
We train the model for $20$ epochs with batch size $64$ via SGD. Refer to Sec. \ref{sec:4.1} and the appendix for other details.

\subsubsection{Results and Discussions}
\label{sec:4.3.2}
\begin{figure}[!t]
\centering
\subfloat{
	\begin{minipage}[t]{0.5\textwidth}
		\flushleft
		\includegraphics[height=1.05in]{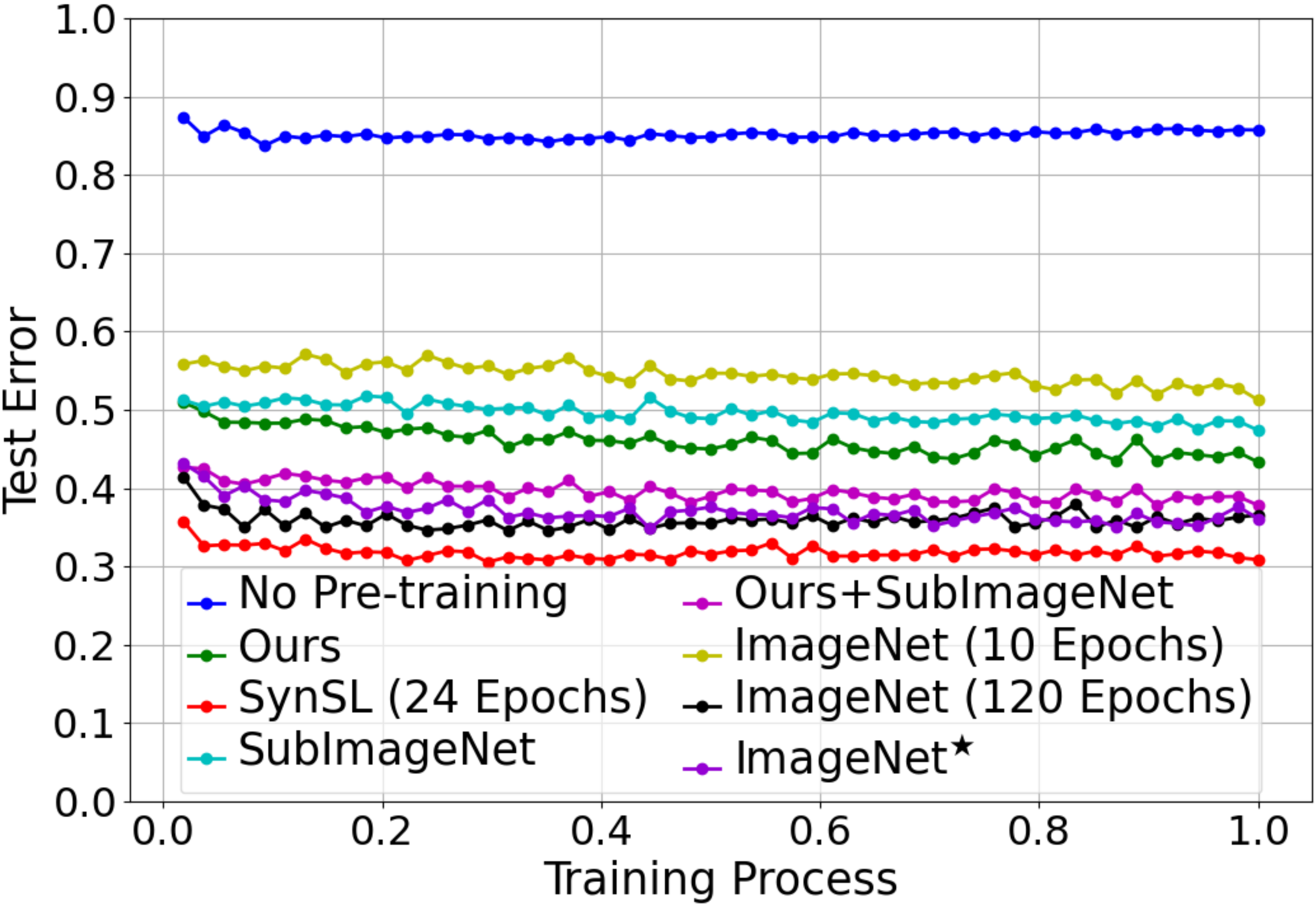}
		\label{fig:da_learning_curves:3}
	\end{minipage}
}
\hspace{-0.45cm}
\subfloat{
	\begin{minipage}[t]{0.5\textwidth}
		\flushleft
		\includegraphics[height=1.05in]{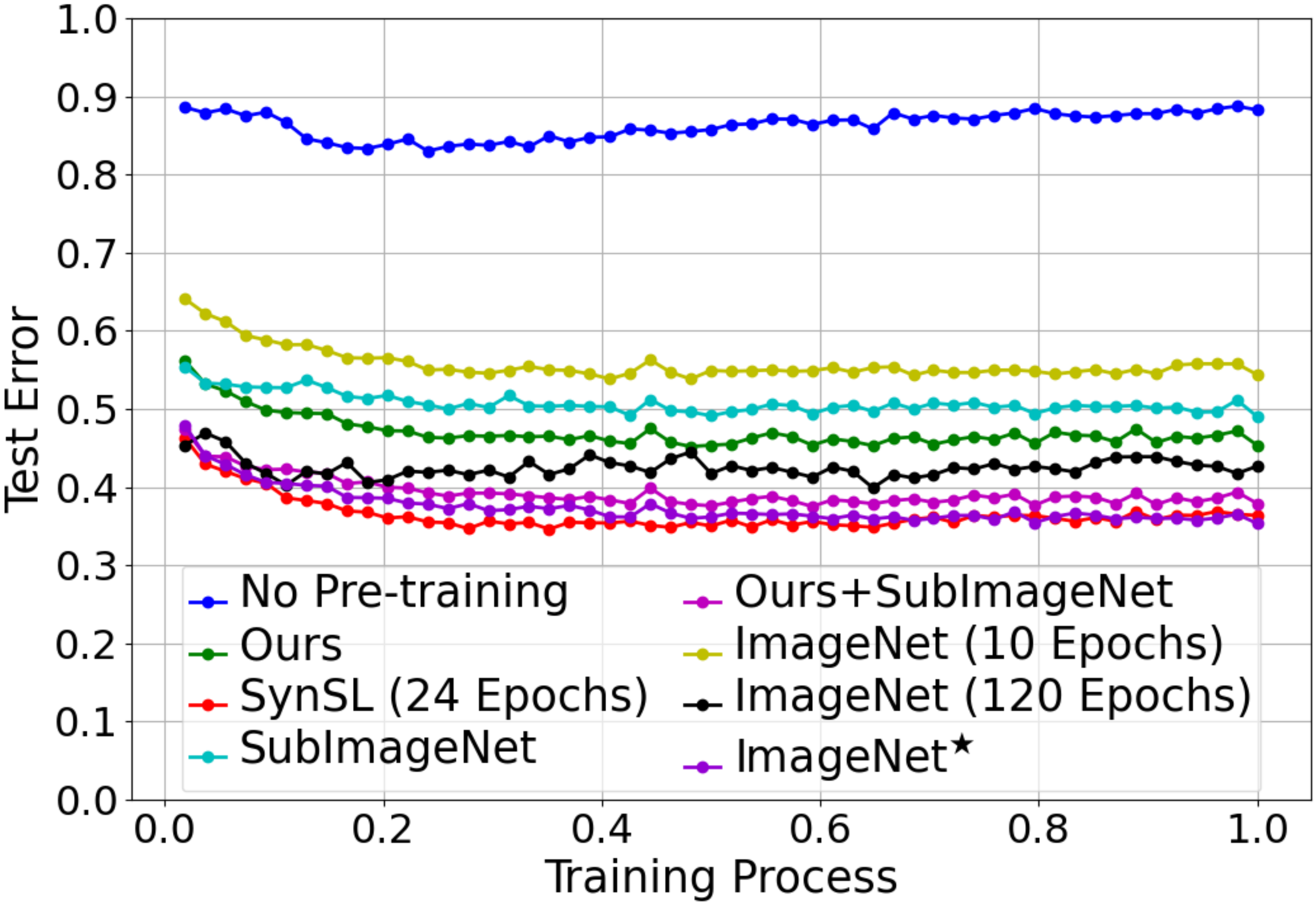}
		\label{fig:da_learning_curves:6}
	\end{minipage}
}
\vspace{-0.3cm}
\caption{Learning process (Mean) of MCD (\textbf{left}) and DisClusterDA (\textbf{right}) when varying the pre-training scheme.}
\label{fig:da_learning_curves}
\end{figure}

To \emph{study the effects of pre-training on synthetic-to-real adaptation}, we examine several DA methods when varying the pre-training scheme in terms of pre-training data and duration. The results are reported in Table \ref{tab:compare_pretrain_data} and Figs. \ref{fig:da_learning_curves} and A7. We emphasize several remarkable findings below.

\noindent\textbf{The importance of pre-training for DA.} DA fails without pre-training. With no pre-training, the very baseline No Adaptation that trains the model only on the labeled source data, outperforms all compared DA methods in overall accuracy, despite the worst mean class precision. It verifies that pre-training is indispensable for DA and involving the target data in training may alleviate class imbalance.

\noindent\textbf{Effects of different pre-training schemes.} Different DA methods exhibit different relative advantages under different pre-training data. When pre-training on our synthesized data, MCD achieves the best results; when pre-training on Ours+SubImageNet, DisClusterDA outperforms the others; 
when pre-training on ImageNet$^\bigstar$, SRDC yields the best performance.
What's worse, the reliability of existing DA method evaluation criteria is unguaranteed. With different pre-training schemes, the best performance is achieved by different DA methods. When pre-training on ImageNet for 10, 60, or 120 epochs, the best results are achieved by RCA, MCD, and SRDC respectively; when pre-training on MetaShift for 5, 30, or 60 epochs, the best results are achieved by RCA, MCD, and DisClusterDA respectively.

\noindent\textbf{Synthetic data pre-training vs. real data pre-training.} Synthetic data pre-training is better than pre-training on real data in our study. With the same 200K pre-training iterations, our synthetic data often bring more benefits than real data from ImageNet or MetaShift, though the top-ranked performance is achieved by extending the pre-training time on real data. 
Under the same experimental configuration, SynSL pre-training for $24$ epochs is comparable to or better than pre-training on ImageNet for 120 epochs and maybe it's because SynSL is 10 times ImageNet's size. The observation indicates that with our 12.8M synthetic data pre-training, the DA methods can yield the new state of the art.

\noindent\textbf{Implications for pre-training data setting.} Big Synthesis Small Real is worth deeply researching. Ours+SubImageNet augmenting our synthetic data with a small amount of real data, achieves remarkable performance gain over Ours, suggesting a promising paradigm of supervised pre-training --- Big Synthesis Small Real.
On the other hand, pre-train with target classes first under limited computing resources. With 200K pre-training iterations, SubImageNet performs much better than ImageNet (10 Epochs), suggesting that one should consider pre-training with target classes first in cases of low computation budget, e.g., real-time deployment on low-power devices like mobile phones, laptops, and smartwatches. Here, we have two questions to be considered: do we have unlimited computing resources for pre-training? Is the domain-specific pre-training more suitable for some industrial applications?

\noindent\textbf{The improved generalization of DA models.} Real data pre-training with extra non-target classes, fine-grained target subclasses, or our synthesized data added for target classes helps DA. 
ImageNet (120 Epochs) involving both target and non-target classes in pre-training is better than SubImageNet involving only target classes, indicating that learning rich category relationships is helpful for downstream transferring.
with 200K pre-training iterations, ImageNet-990 performs much worse than ImageNet, implying that pre-training in a fine-grained visual categorization manner may bring surprising benefits.
Ours+SubImageNet adding our synthesized data for target classes in SubImageNet, produces significant improvements and is close to ImageNet (120 Epochs); ImageNet-990+Ours improves over ImageNet-990, suggesting that synthetic data may help improve the performance further.

\noindent\textbf{Convergence analysis.} In Figs. \ref{fig:da_learning_curves} and A7, the convergence from different pre-training schemes for the same DA method differs in speed, stability, and accuracy. In \cref{fig:da_learning_curves}, SynSL with $24$ epochs outperforms ImageNet with $120$ epochs significantly; notably, SynSL is on par with or better than ImageNet$^\bigstar$, supporting our aforementioned findings.

\begin{figure}[!t]
\centering
\includegraphics[width=1.0\textwidth]{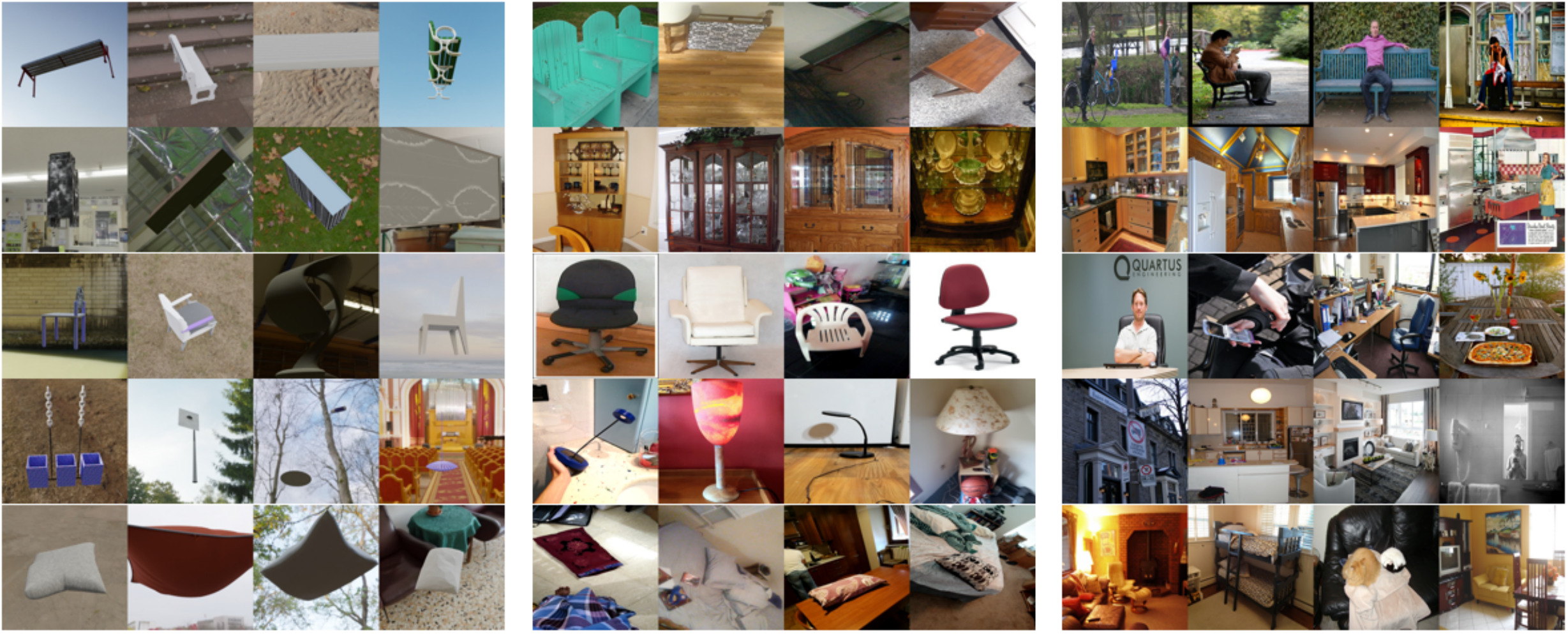}
\vspace{-0.5cm}
\caption{Sample images from the synthetic (left) domain and the real domains of S2RDA-49 (middle) and S2RDA-MS-39 (right).}
\label{fig:new_benchmark}
\end{figure}

\begin{figure*}[!t]
\centering
\subfloat[No Adaptation]{
	\begin{minipage}[t]{0.25\textwidth}
		\centering
		\includegraphics[height=1.5in]{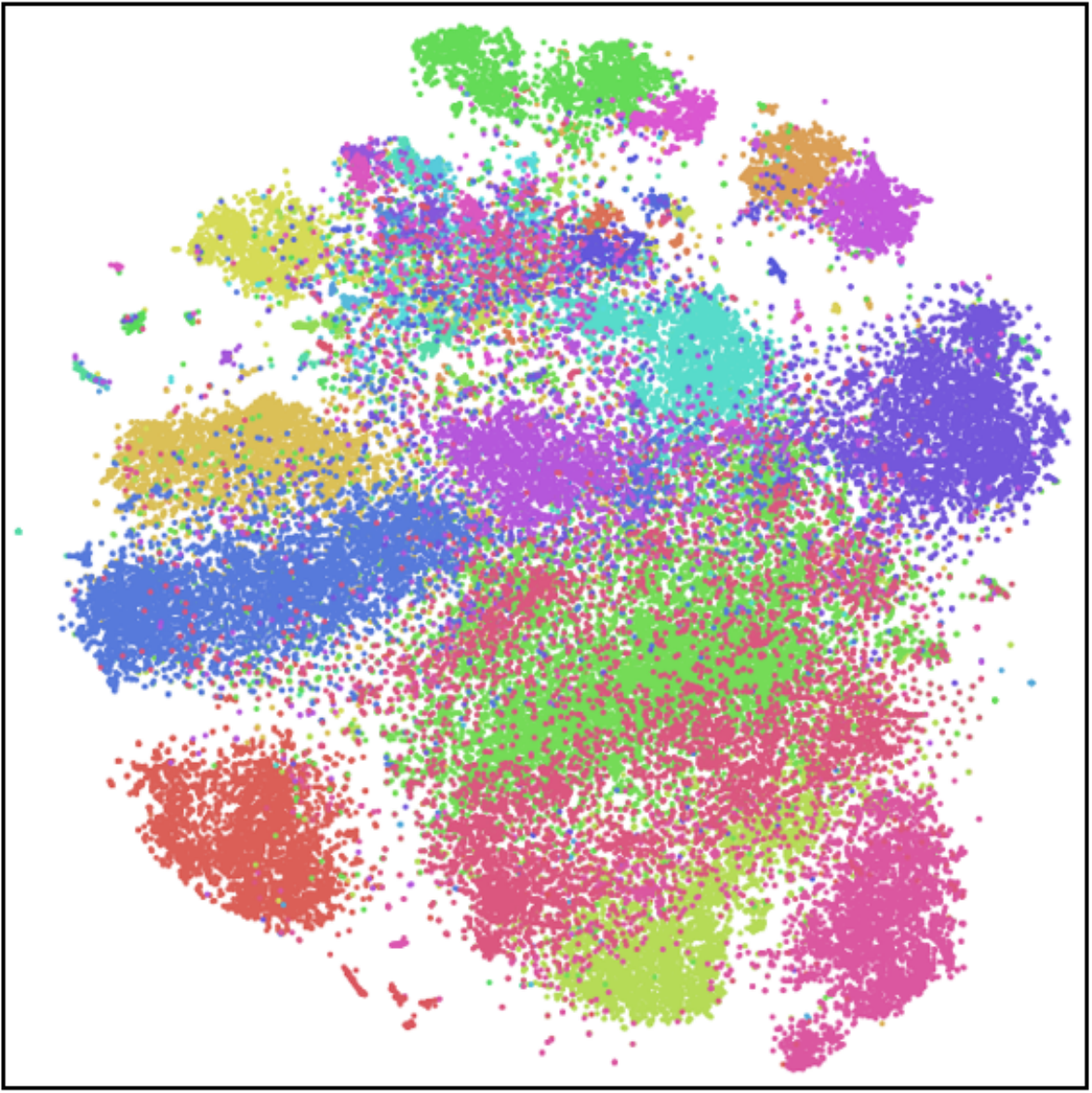}
		\label{fig:tsne:1}
	\end{minipage}
}
\subfloat[SRDC]{
	\begin{minipage}[t]{0.25\textwidth}
		\centering
		\includegraphics[height=1.5in]{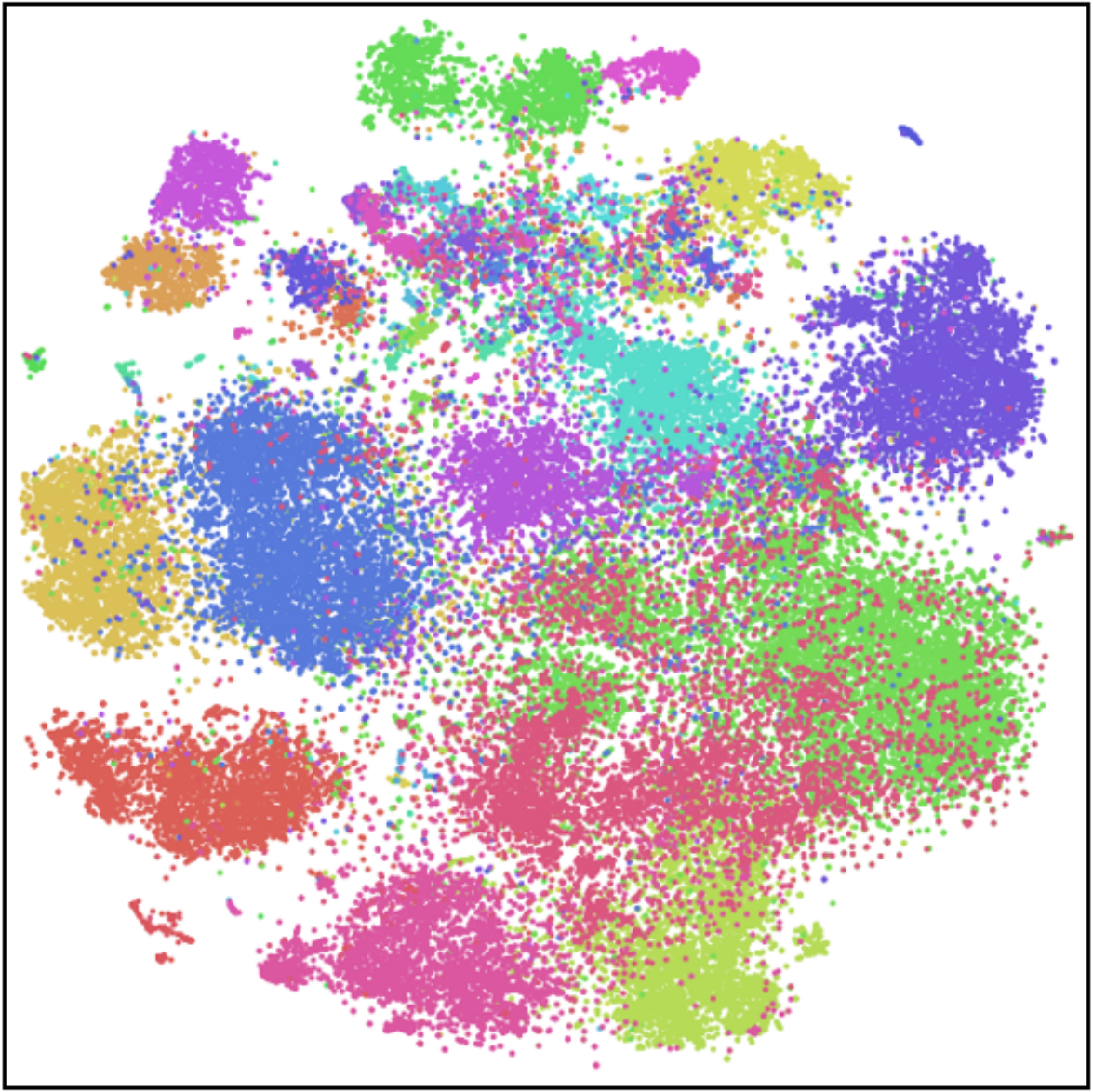}
		\label{fig:tsne:2}
	\end{minipage}
}
\subfloat[No Adaptation]{
	\begin{minipage}[t]{0.25\textwidth}
		\centering
		\includegraphics[height=1.5in]{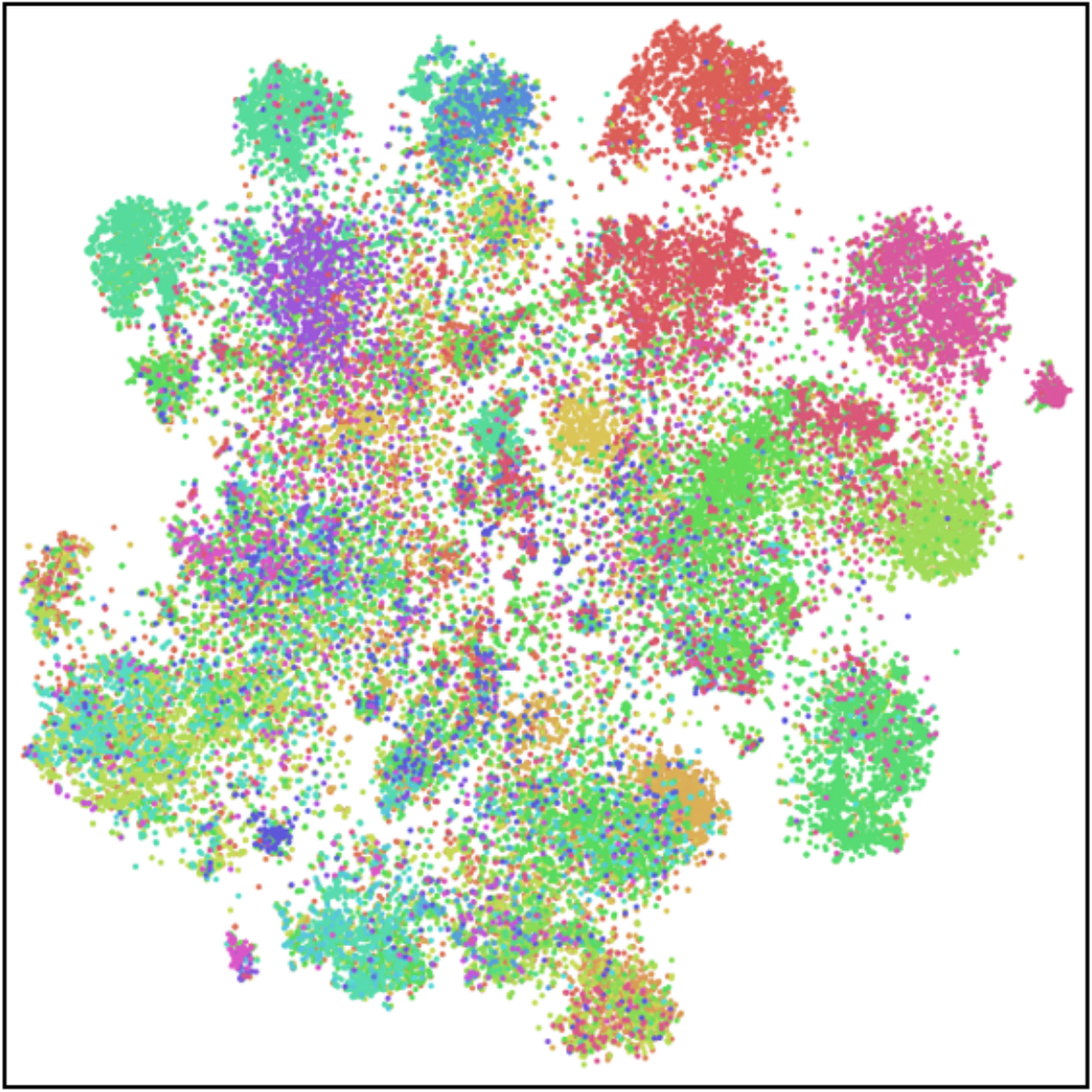}
		\label{fig:tsne:3}
	\end{minipage}
}
\subfloat[DisClusterDA]{
	\begin{minipage}[t]{0.25\textwidth}
		\centering
		\includegraphics[height=1.5in]{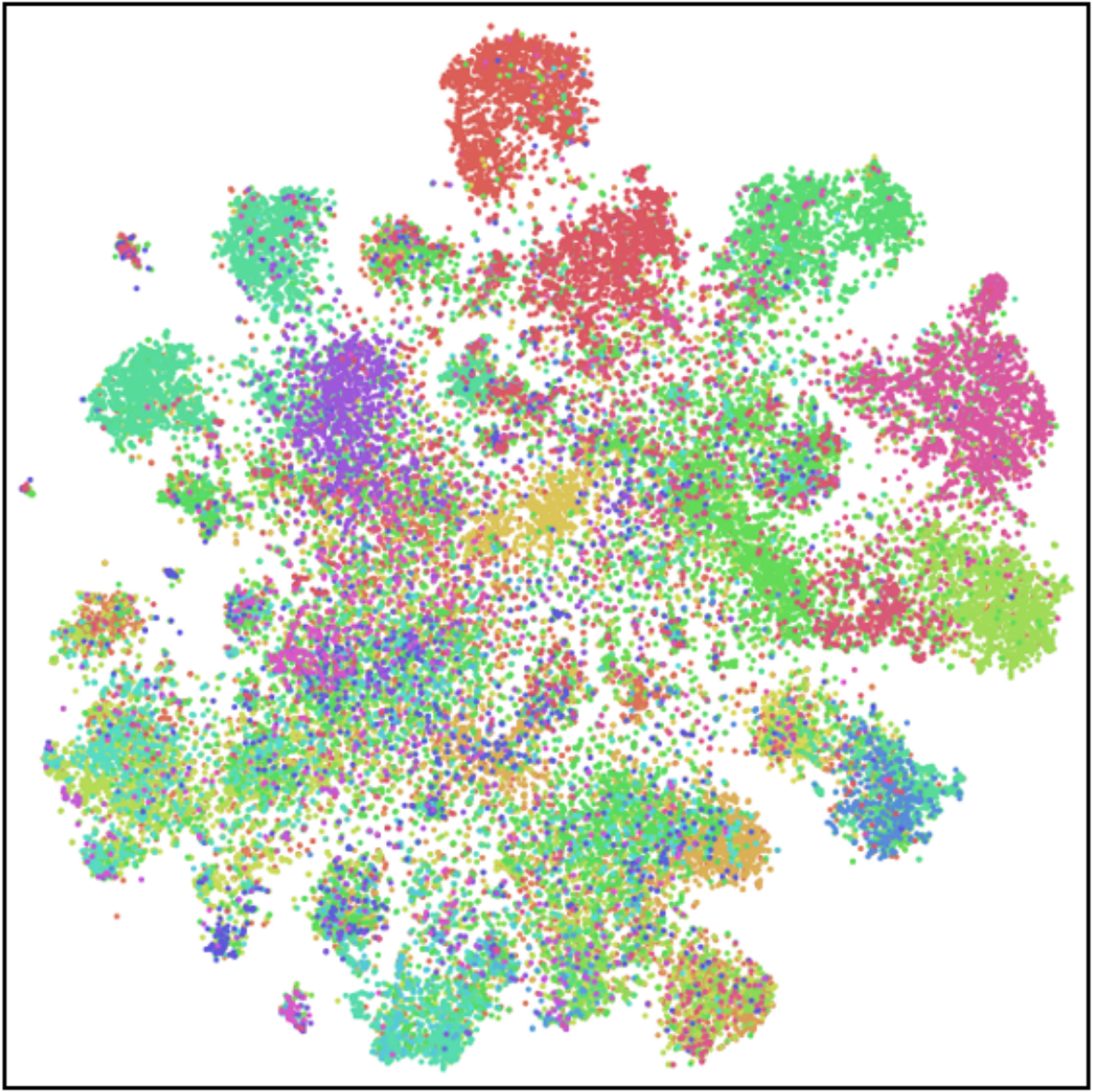}
		\label{fig:tsne:4}
	\end{minipage}
}
\vspace{-0.3cm}
\caption{The t-SNE visualization of target domain features extracted by different models on S2RDA-49 (a-b) and S2RDA-MS-39 (c-d).}
\label{fig:tsne}
\end{figure*}

\subsection{A New Synthetic-to-Real Benchmark}
\label{sec:4.4}

\noindent\textbf{Dataset Construction.} Our proposed Synthetic-to-Real benchmark for more practical visual DA (termed S2RDA) includes two challenging transfer tasks of S2RDA-49 and S2RDA-MS-39 (cf. Fig. \ref{fig:new_benchmark}). In each task, source/synthetic domain samples are synthesized by rendering 3D models from ShapeNet \cite{shapenet2015} (cf. Sec. \ref{sec:renderer})
. The used 3D models are in the same label space as the target/real domain and each class has $12$K rendered RGB images. 
The real domain of S2RDA-49 comprises $60,535$ images of $49$ classes, collected from ImageNet validation set, ObjectNet \cite{ObjectNet}, VisDA-2017 validation set, and the web \cite{kaggle}
. 
For S2RDA-MS-39, the real domain collects $41,735$ natural images exclusive for $39$ classes from MetaShift \cite{MetaShift}, which contain complex and distinct contexts, e.g., object presence (co-occurrence of different objects), general contexts (indoor or outdoor), and object attributes (color or shape), leading to a much harder task. 
In Fig. A8, we show the long-tailed distribution of image number per class in each real domain
. 
Compared to VisDA-2017 \cite{visda2017}, our S2RDA contains more categories, more realistically synthesized source domain data coming for free, and more complicated target domain data collected from diverse real-world sources, setting a more practical, challenging benchmark for future DA research.

\begin{table}[!t]
	\caption{Domain adaptation performance on S2RDA (ResNet-50).}
	\vspace{-0.3cm}
	\label{tab:new_benchmark}
	\centering
	\resizebox{1.0\linewidth}{!}{
		\begin{tabular}{l >{\columncolor[gray]{0.78}}c c >{\columncolor[gray]{0.78}}c c >{\columncolor[gray]{0.78}}c c >{\columncolor[gray]{0.78}}c c >{\columncolor[gray]{0.78}}c c >{\columncolor[gray]{0.78}}c c}
		\toprule
		\multirow{2}{*}{Transfer Task} & \multicolumn{2}{c}{No Adaptation} & \multicolumn{2}{c}{DANN} & \multicolumn{2}{c}{MCD} & \multicolumn{2}{c}{RCA} & \multicolumn{2}{c}{SRDC} & \multicolumn{2}{c}{DisClusterDA} \\
		& Acc. & Mean & Acc. & Mean & Acc. & Mean & Acc. & Mean & Acc. & Mean & Acc. & Mean \\
		\midrule
		S2RDA-49 & 51.89 & 42.19 & 47.06 & 47.64 & 42.51 & 47.77 & 47.07 & 48.46 & \hl{\textbf{61.52}} & \ul{\textbf{52.98}} & 53.03 & 52.34 \\
		
		S2RDA-MS-39 & 22.03 & 20.54 & 22.82 & 22.20 & 22.07 & 22.16 & 23.34 & 22.53 & 25.83 & 24.55 & \hl{\textbf{27.14}} & \ul{\textbf{25.33}} \\
		
		\bottomrule
	\end{tabular}
}\vspace{-0.3cm}
\end{table}

\noindent\textbf{Benchmarking DA Methods.} We use ResNet-50 as the backbone, initialized by the official ImageNet pre-trained checkpoint \cite{resnet}. 
Refer to Sec. \ref{sec:4.3} and the appendix for other implementation details. 
We report the results on S2RDA in Table \ref{tab:new_benchmark} and show t-SNE \cite{tsne} visualizations in Fig. \ref{fig:tsne}. In the overall accuracy (Acc.) on S2RDA-49, the adversarial training based methods DANN, MCD, and RCA perform worse than No Adaptation, demonstrating that explicit domain adaptation could deteriorate the intrinsic discriminative structures of target domain data \cite{srdc,disclusterda}, especially in cases of more categories; in contrast, SRDC produces significant quantitative and qualitative improvements
, $14.45\%$ and $8.49\%$ higher than RCA and DisClusterDA respectively, but $7.48\%$ lower than Acc. on SubVisDA-10 (cf. Table \ref{tab:compare_pretrain_data}), indicating the difficulty of the S2RDA-49 task. On S2RDA-MS-39, the classification accuracy is much worse than that on S2RDA-49 (decrease of more than $20\%$), and the compared methods show much less performance difference; the highest accuracy is only $27.14\%$ achieved by DisClusterDA, showing that S2RDA-MS-39 is a very challenging task. To summarize, \emph{domain adaptation is far from being solved} and we expect that our results contribute to the DA community as new benchmarks, though more careful studies in different algorithmic frameworks are certainly necessary to be conducted.

\section{Conclusions and Future Perspectives}
\label{sec:conclusion}
This paper primarily aims to publish new datasets including our synthetic dataset SynSL (12.8M) and S2RDA, and benchmarks the datasets via supervised learning and downstream transferring. 
In the context of image classification, our work is the first comprehensive study on synthetic data learning, which is completely missing. 
We propose exploiting synthetic datasets to explore questions on model generalization 
and benchmark pre-training strategies for DA. We build randomized synthetic datasets using a 3D rendering engine and use this established system to manipulate the generation of images by altering several imaging factors
. We find that synthetic data pre-training has the potential to be better than pre-training on real data, 
our new benchmark S2RDA is much more practical for synthetic-to-real DA, to name a few. We expect that these results contribute to the transfer learning community as new benchmarks, though the research on more synthetic datasets, more models, and more DA methods is certainly to be done.


\noindent\emph{Synthetic data as a new benchmark.} Synthetic data are well suited for use as toy examples to verify existing deep learning theoretical results or explore new theories. 

\noindent\emph{Evaluation metrics robust to pre-training.} The comparison among various DA methods yields different or even opposite results when using different pre-training schemes (cf. \cref{sec:4.3}). DA researchers should propose and follow evaluation metrics enabling effective and fair comparison.

\noindent\emph{More realistic simulation synthesis.} We will consider more imaging parameters, e.g., randomizing the type and hue of the light, including 77 physical objects with actual textures from YCB \cite{YCB}, and using the flying distractor \cite{SIDOD}.

\noindent\emph{To explore deep learning based data generation.} Our proposed paradigm of empirical study can generalize to any data generation pipeline. Our findings may be data source specific and the generalizability to other pipelines like GANs, NeRFs, and AutoSimulate \cite{AutoSimulate} is to be explored. 

\noindent\emph{Applicability to other vision tasks.} 
Our new paradigm of empirical study for image classification can also be applied to other vision tasks of semantic analysis, e.g., Kubric \cite{Kubric} and HyperSim \cite{hypersim} for segmentation and object detection.

\noindent\textbf{Acknowledgments.} This work is supported in part by Program for Guangdong Introducing Innovative and Enterpreneurial Teams (No.: 2017ZT07X183) and Guangdong R\&D key project of China (No.: 2019B010155001).

{\small
\bibliographystyle{ieee_fullname}
\bibliography{egbib}
}

\newpage

\begin{appendices}
\renewcommand{\thefigure}{A\arabic{figure}}
\renewcommand{\thetable}{A\arabic{table}}
\setcounter{figure}{0}
\setcounter{table}{0}

\twocolumn[{%
	\centering
	\LARGE Appendix for \\ A New Benchmark: On the Utility of Synthetic Data with Blender for \\ Bare Supervised Learning and Downstream Domain Adaptation\\[2.5em]
}]



The catalog of this appendix is in the following.
\begin{itemize}
	\item Sec. \ref{sec:contributions} summarizes our main contributions.
	
	\item Sec. \ref{sec:paper_novelty} provides a well-organized summary for the paper novelty.
	
	\item Sec. \ref{sec:more_clarifications} makes more clarifications on our empirical study.
	
	
	
	\item Sec. \ref{sec:learning_process} examines the learning process, Sec. \ref{sec:saliency_maps} visualizes the saliency map, and Sec. \ref{sec:impact_data_aug} depicts more impact of data augmentations for the comprehensive comparison between fixed-dataset periodic training and training on non-repetitive samples.
	
	\item Sec. \ref{sec:compare_3_models} evaluates various network architectures by plotting their learning curves.
	
	\item Sec. \ref{sec:compare_pretrain_for_da} shows the learning process of different pre-training data using domain adaptation as the downstream task.
	
	\item Sec. \ref{sec:s2r} presents more details on our proposed S2RDA benchmark, e.g., comparing synthetic data with real data from different angles.
	
	\item Sec. \ref{sec:other_details} provides other implementation details for supervised learning/pre-training and downstream domain adaptation.
	
	\item Sec. \ref{sec:related_works} reviews other related works on real datasets, data manipulation \cite{survey_dg,survey_data_aug,domain_randomization}, deep models \cite{survey_deep_cnn,ViT,MLP-Mixer}, transfer learning \cite{survey_tl_early,survey_tl_recent,survey_transferability_in_dl}, domain adaptation \cite{survey_da}, and OOD generalization \cite{survey_dg}.
\end{itemize}

\section{Our Main Contributions}
\label{sec:contributions}

Our main contributions are summarized as follows.
\begin{itemize}
	\item On the well-controlled IID experimental condition enabled by 3D rendering, we empirically verify the typical insights on shortcut learning, PAC generalization, and variance-bias trade-off, and explore the effects of changing data regimes and network structures on model generalization. The key design wherein is to compare the traditional fixed-dataset periodic training with a new strategy of training on non-repetitive samples.
	
	\item We explore how variation factors of an image affect the model generalization, e.g., object scale, material texture, illumination, camera viewpoint, and background, and in return provide new perceptions for data generation.
	
	\item Using the popular simulation-to-real classification adaptation as a downstream task, we investigate how synthetic data pre-training performs by comparing with pre-training on real data. We have some surprising and important discoveries including synthetic data pre-training is also prospective and a promising paradigm of pre-training on big synthetic data together with small real data is proposed for realistic supervised pre-training.
	
	\item We propose a more large-scale synthetic-to-real benchmark for classification adaptation (termed S2RDA), on which we also provide a baseline performance analysis for representative DA approaches.
\end{itemize}

\section{Summary of Paper Novelty}
\label{sec:paper_novelty}

Now it is becoming more and more important to work on methods that use simulated data but perform well in practical domains whose data or annotation are difficult to acquire, e.g., medical imaging. However, previous research works \emph{have not} studied various factors on a synthesized dataset for image classification and domain adaptation comprehensively and systematically. To fill the gap, we present \emph{the first work}, ranging from bare supervised learning to downstream domain adaptation. It provides many new, valuable learning insights for OOD/real data generalization, though the verification of some existing, known theories in our well-controlled IID experimental condition has also been done for comprehensive coverage. It is essential for synthetic data learning analysis, which is \emph{completely missing} in the context of image classification. We clarify the paper novelty below.

\begin{itemize}
	\item The motivation that we utilize synthetic data to verify typical theories and expose new findings is novel. Real data are noisy and uncontrolled, which may hinder the verification of typical theories and exposure to new findings. In the context of image classification, existing works verify classical theories and reveal new findings on real data. However, the process of acquiring real data cannot be controlled, the annotation accuracy cannot be guaranteed, and there may be duplicate images in the training set and test set, which leads to the fact that the training set and test set are no longer independent and identically distributed (IID). To remedy them, we resort to synthetic data generated by 3D rendering with domain randomization.
	
	\item The comparison between fixed-dataset periodic training and training on non-repetitive samples and the study of shortcut learning on our synthesized dataset are novel. We admit that some of our findings are classical theories, e.g., PAC generalization and variance-bias trade-off, which should be verified when one introduces a new dataset. We introduce a new dataset of synthetic data and thus do such a study for comprehensive coverage, which first compares fixed-dataset periodic training with training on non-repetitive samples generated by 3D rendering. Particularly, we also verify a recent, significant perspective of shortcut learning and design new experiments to demonstrate that randomizing the variation factors of training images can block shortcut solutions that rely on context clues in the background.
	
	\item Investigating the learning characteristics and properties of our synthesized new dataset comprehensively is novel, and our experiments yield many interesting and valuable observations. Synthetic data are cheap, label-rich, and well-controlled, but there hasn’t been a comprehensive study of bare supervised learning on synthetic data in the context of image classification. To our knowledge, we are the first to investigate the learning characteristics and properties of our synthesized new dataset comprehensively, in terms of refreshed architecture, model capacity, training data quantity, data augmentation, and rendering variations. The empirical study on bare supervised learning yields many new findings, e.g.,
	\begin{itemize}
		\item IID and OOD generalizations are some type of zero-sum game,
		\item ViT performs surprisingly poorly,
		\item there is always a bottleneck from synthetic data to OOD/real data,
		\item neural architecture search (NAS) should also consider the search for data augmentation, and
		\item different factors and even their different values have uneven importance to IID generalization.
	\end{itemize}
	
	\item Synthetic data pre-training, its comparison to real data pre-training, and its application to downstream synthetic-to-real classification adaptation are novel, and our experiments yield many interesting and valuable observations. To our knowledge, there is little research on pre-training for domain adaptation. Kin et al. \cite{study_pretrain_for_DA} preliminarily study the effects of real data pre-training on domain transfer tasks. Differently, we focus on the learning utility of synthetic data and take the first step towards clearing the cloud of mystery surrounding how different pre-training schemes including synthetic data pre-training affect the practical, large-scale synthetic-to-real classification adaptation. Besides, we first study and compare pre-training on the latest MetaShift dataset \cite{MetaShift}. The empirical study on downstream domain adaptation yields many new findings, e.g.,
	\begin{itemize}
		\item DA fails without pre-training,
		\item different DA methods exhibit different relative advantages under different pre-training data,
		\item the reliability of existing DA method evaluation criteria is unguaranteed,
		\item synthetic data pre-training is better than pre-training on real data (e.g., ImageNet) in our study, and
		\item Big Synthesis Small Real is worth deeply researching.
	\end{itemize}
	
	\item Our introduced S2RDA benchmark is novel and will advance the field of domain adaptation research on transfer from synthetic to real.
	
\end{itemize}

Our findings may challenge some of the current conclusions, but they also shed some light on the important fields in computer vision and take a step towards uncovering the mystery of deep learning.

\begin{table*}[t]
	\caption{Domain adaptation performance on SubVisDA-10 with varied pre-training schemes (ResNet-50). 
		Green or red: Best Acc. or Mean in each row (among compared DA methods).}
	\label{tab:study_imagenet990}
	\centering
	\resizebox{0.95\linewidth}{!}{
		\begin{tabular}{lcc >{\columncolor[gray]{0.78}}c c >{\columncolor[gray]{0.78}}c c >{\columncolor[gray]{0.78}}c c >{\columncolor[gray]{0.78}}c c >{\columncolor[gray]{0.78}}c c >{\columncolor[gray]{0.78}}c c}
			\toprule
			\multirow{2}{*}{Pre-training Data} & \multirow{2}{*}{\# Iters} & \multirow{2}{*}{\# Epochs} & \multicolumn{2}{c}{No Adaptation} & \multicolumn{2}{c}{DANN} & \multicolumn{2}{c}{MCD} & \multicolumn{2}{c}{RCA} & \multicolumn{2}{c}{SRDC} & \multicolumn{2}{c}{DisClusterDA} \\
			& & & Acc. & Mean & Acc. & Mean & Acc. & Mean & Acc. & Mean & Acc. & Mean & Acc. & Mean \\
			\midrule
			
			ImageNet-990 & 200K & 10 & 50.11 & 45.45 & 55.68 & 54.67 & 58.84 & 58.44 & 58.79 & \ul{60.18} & \hl{60.25} & 57.68 & 57.62 & 57.42 \\
			
			ImageNet-990+Ours & 200K & 9 & 52.87 & \textbf{48.85} & \textbf{58.42} & \textbf{58.02} & 60.52 & \textbf{62.27} & \textbf{62.28} & \textbf{\ul{63.35}} & \textbf{\hl{63.60}} & \textbf{60.89} & \textbf{61.90} & \textbf{63.10} \\
			
			\midrule
			
			ImageNet & 200K & 10 & \textbf{53.24} & 45.38 & 57.77 & 55.59 & \textbf{61.90} & \ul{61.75} & 61.59 & 60.72 & \hl{62.56} & 59.24 & 61.18 & 59.01 \\

			\bottomrule
		\end{tabular}
	}
\end{table*}

\section{More Clarifications on Our Empirical Study}
\label{sec:more_clarifications}

\paragraph{Necessity of domain randomization study in Table 2.} 
We expect that assessing variation factors should be essential for synthesizing data for image classification, which is \emph{missing} in previous work \cite{visda2017}. We admit that such a study has been done for detection and scene understanding in prior methods \cite{domain_randomization,SDR4CarDetection,DR4ObjectDetection}, but the generalizability of those results to image classification is \emph{lack of guarantee}. Thus, we follow them and do the study for image classification, where we consider a \emph{different} set of factors (e.g., light, texture, and flying distractors in \cite{DR4ObjectDetection}). It may be expected that a fixed value would underperform randomized values, but how much each factor and each value degrade is \emph{unknown}. One cannot know how important they are without such a \emph{quantitative} study. Insightfully, our new results in Table 2 also suggest that the under-explored direction of weighted rendering is worth studying and provide preliminary guidance/prior knowledge for learning factor distribution.

\paragraph{Additonal experiments on ImageNet-990+Ours improving over ImageNet-990.}
To further justify that ImageNet-990+Ours improves over ImageNet-990, we do additional experiments by using a different cosine decay schedule for the learning rate: $\eta_p = \eta_1 + 0.5 (\eta_0 - \eta_1) (1 + \cos (\pi p))$, where $p$ is the process of training iterations normalized to be in $[0, 1]$, the initial learning rate $\eta_0 = 0.1$, and the final learning rate $\eta_1 = 0.001$. The results for several DA methods are reported in Table \ref{tab:study_imagenet990}. As we can see, with fine-grained subclasses merged into one, ImageNet-990 underperforms ImageNet by a large margin, suggesting that it may be helpful to use fine-grained visual categorization for pre-training; in contrast, by adding our 120K synthetic images, ImageNet-990+Ours is comparable to or better than ImageNet, confirming the utility of our synthetic data.

\paragraph{Pre-training with an increased number of classes helps DA.} In Table 3, we have compared SubImageNet involving only target classes in training with ImageNet involving both target and non-target classes; with abundant pre-training epochs, the latter is evidently better than the former, indicating that learning rich category relationships is helpful for downstream DA. \emph{A similar phenomenon is observed in synthetic data pre-training.} For example, we have done experiments of pre-training on the synthetic domain of our proposed S2RDA-49 task; compared with pre-training on Ours (120K images, 10 classes), MCD improves by 5.97\% in Acc. and 6.06\% in Mean, and DisClusterDA improves by 4.69\% in Acc. and 5.36\% in Mean.

\paragraph{Necessity and applicability of the proposed S2RDA.}
Note that SRDC outperforms the baseline No Adaptation by $\sim$10\% on S2RDA-49 and DisClusterDA outperforms that by $\sim$5\%, verifying the efficacy of these DA methods. The observations also demonstrate that S2RDA \emph{can} benchmark different DA methods. Compared to SubVisDA-10 (cf. Table 3), SRDC degrades by $\sim$7\% on S2RDA-49, which is reasonable as our real domain contains more practical images from real-world sources, though \emph{our synthetic data contain much more diversity, e.g., background} (cf. Fig. 1).
Differently, S2RDA-MS-39, which decreases by $>$20\% over S2RDA-49, evaluates different DA approaches on the \emph{worst/extreme cases} (cf. Fig. 6), making a more comprehensive comparison and acting as a touchstone to examine and advance DA algorithms.
Reducing the domain gap between simple and difficult backgrounds is by nature one of the key issues in simulation-to-real transfer, as also shown in \cite{visda2017}; therefore, reducing such a domain gap is one of the criteria for judging excellent DA methods.
To sum up, S2RDA is a \emph{better} benchmark than VisDA-2017, as it has more realistic synthetic data and more practical real data with more object categories, and enables a larger room of improvement for promoting the progress of DA algorithms and models.

%

%
%

\section{Fixed-Dataset Periodic Training vs. Training on Non-Repetitive Samples}
\label{sec:fix_vs_unfix}

\subsection{Examining Learning Process}
\label{sec:learning_process}

\begin{figure*}[!ht]
	\centering
	\subfloat[Training loss]{
		\begin{minipage}[t]{0.25\textwidth}
			\includegraphics[height=0.85in]{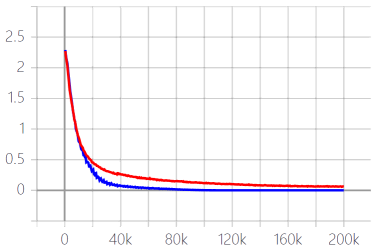}
			\label{fig:learning_curves_all:1}
		\end{minipage}
	}
	\subfloat[Test loss (IID)]{
		\begin{minipage}[t]{0.25\textwidth}
			\includegraphics[height=0.85in]{images/learning_curves/test_1.test_loss_da-n_res50_syn_id.svg}
			\label{fig:learning_curves_all:2}
		\end{minipage}
	}
	\subfloat[Training acc.]{
		\begin{minipage}[t]{0.25\textwidth}
			\includegraphics[height=0.85in]{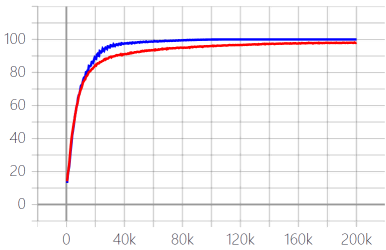}
			\label{fig:learning_curves_all:3}
		\end{minipage}
	}
	\subfloat[Test acc. (IID)]{
		\begin{minipage}[t]{0.26\textwidth}
			\includegraphics[height=0.85in]{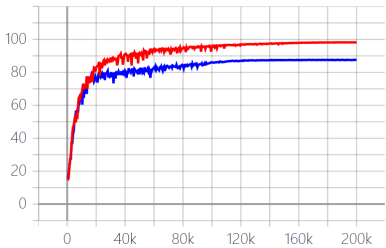}
			\label{fig:learning_curves_all:4}
		\end{minipage}
	}
	\\
	\subfloat[Test loss (IID w/o BG)]{
		\begin{minipage}[t]{0.25\textwidth}
			\includegraphics[height=0.85in]{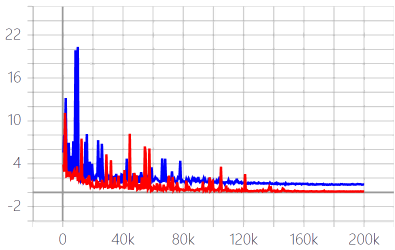}
			\label{fig:learning_curves_all:5}
		\end{minipage}
	}
	\subfloat[Test loss (OOD)]{
		\begin{minipage}[t]{0.25\textwidth}
			\includegraphics[height=0.85in]{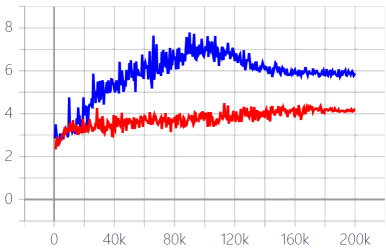}
			\label{fig:learning_curves_all:6}
		\end{minipage}
	}
	\subfloat[Test acc. (IID w/o BG)]{
		\begin{minipage}[t]{0.25\textwidth}
			\includegraphics[height=0.85in]{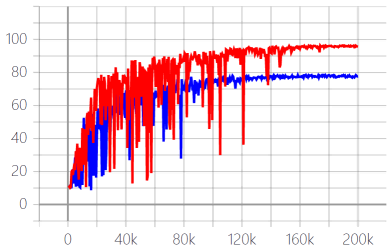}
			\label{fig:learning_curves_all:7}
		\end{minipage}
	}
	\subfloat[Test acc. (OOD)]{
		\begin{minipage}[t]{0.26\textwidth}
			\includegraphics[height=0.85in]{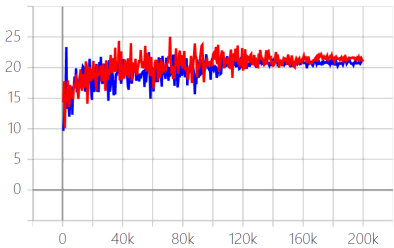}
			\label{fig:learning_curves_all:8}
		\end{minipage}
	}
	\\
	\subfloat[Training loss]{
		\begin{minipage}[t]{0.25\textwidth}
			\includegraphics[height=0.85in]{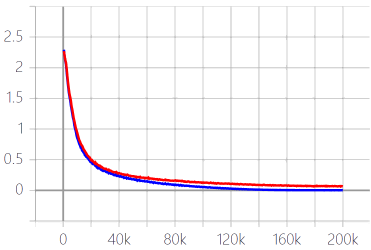}
			\label{fig:learning_curves_all:9}
		\end{minipage}
	}
	\subfloat[Test loss (IID)]{
		\begin{minipage}[t]{0.25\textwidth}
			\includegraphics[height=0.85in]{images/learning_curves/test_1.test_loss_da-w_res50_syn_id.svg}
			\label{fig:learning_curves_all:10}
		\end{minipage}
	}
	\subfloat[Training acc.]{
		\begin{minipage}[t]{0.25\textwidth}
			\includegraphics[height=0.85in]{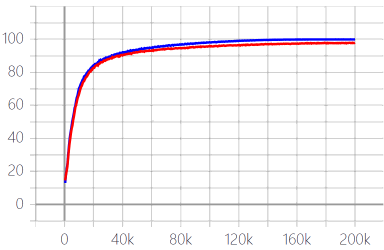}
			\label{fig:learning_curves_all:11}
		\end{minipage}
	}
	\subfloat[Test acc. (IID)]{
		\begin{minipage}[t]{0.26\textwidth}
			\includegraphics[height=0.85in]{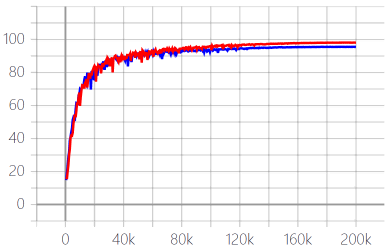}
			\label{fig:learning_curves_all:12}
		\end{minipage}
	}
	\\
	\subfloat[Test loss (IID w/o BG)]{
		\begin{minipage}[t]{0.25\textwidth}
			\includegraphics[height=0.85in]{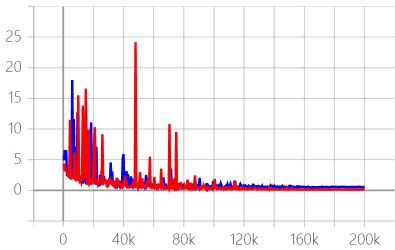}
			\label{fig:learning_curves_all:13}
		\end{minipage}
	}
	\subfloat[Test loss (OOD)]{
		\begin{minipage}[t]{0.25\textwidth}
			\includegraphics[height=0.85in]{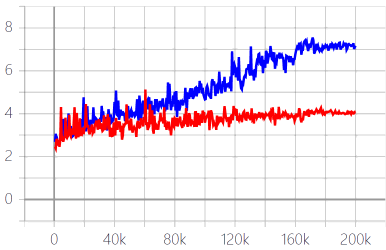}
			\label{fig:learning_curves_all:14}
		\end{minipage}
	}
	\subfloat[Test acc. (IID w/o BG)]{
		\begin{minipage}[t]{0.25\textwidth}
			\includegraphics[height=0.85in]{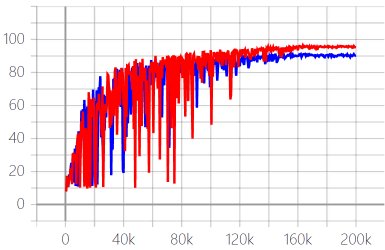}
			\label{fig:learning_curves_all:15}
		\end{minipage}
	}
	\subfloat[Test acc. (OOD)]{
		\begin{minipage}[t]{0.26\textwidth}
			\includegraphics[height=0.85in]{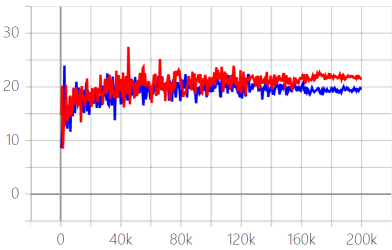}
			\label{fig:learning_curves_all:16}
		\end{minipage}
	}
	\\
	\subfloat[Training loss]{
		\begin{minipage}[t]{0.25\textwidth}
			\includegraphics[height=0.85in]{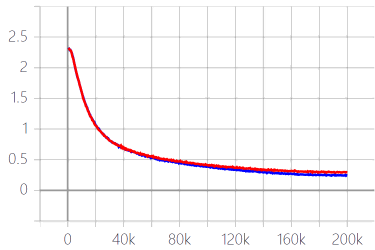}
			\label{fig:learning_curves_all:17}
		\end{minipage}
	}
	\subfloat[Test loss (IID)]{
		\begin{minipage}[t]{0.25\textwidth}
			\includegraphics[height=0.85in]{images/learning_curves/test_1.test_loss_da-s_res50_syn_id.svg}
			\label{fig:learning_curves_all:18}
		\end{minipage}
	}
	\subfloat[Training acc.]{
		\begin{minipage}[t]{0.25\textwidth}
			\includegraphics[height=0.85in]{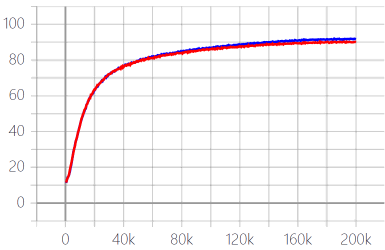}
			\label{fig:learning_curves_all:19}
		\end{minipage}
	}
	\subfloat[Test acc. (IID)]{
		\begin{minipage}[t]{0.26\textwidth}
			\includegraphics[height=0.85in]{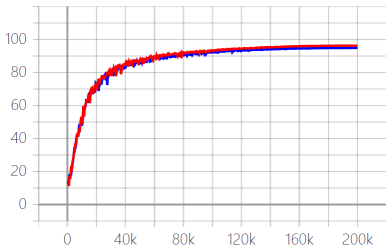}
			\label{fig:learning_curves_all:20}
		\end{minipage}
	}
	\\
	\subfloat[Test loss (IID w/o BG)]{
		\begin{minipage}[t]{0.25\textwidth}
			\includegraphics[height=0.85in]{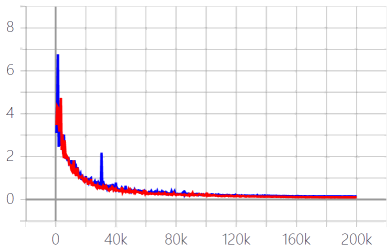}
			\label{fig:learning_curves_all:21}
		\end{minipage}
	}
	\subfloat[Test loss (OOD)]{
		\begin{minipage}[t]{0.25\textwidth}
			\includegraphics[height=0.85in]{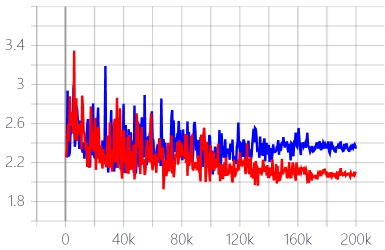}
			\label{fig:learning_curves_all:22}
		\end{minipage}
	}
	\subfloat[Test acc. (IID w/o BG)]{
		\begin{minipage}[t]{0.25\textwidth}
			\includegraphics[height=0.85in]{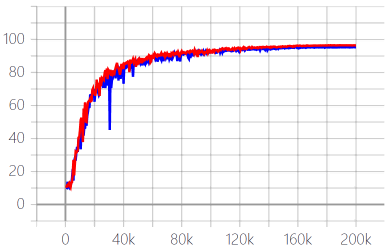}
			\label{fig:learning_curves_all:23}
		\end{minipage}
	}
	\subfloat[Test acc. (OOD)]{
		\begin{minipage}[t]{0.26\textwidth}
			\includegraphics[height=0.85in]{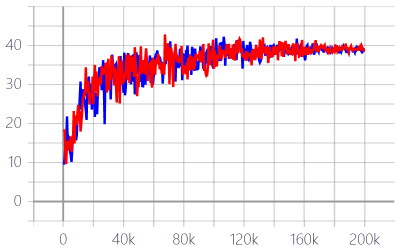}
			\label{fig:learning_curves_all:24}
		\end{minipage}
	}
	\caption{Learning process of training ResNet-50 on a fixed dataset (\textcolor{blue}{blue}) or non-repetitive samples (\textcolor{red}{red}). Note that (a-h), (i-p), and (q-x) are for no, weak, and strong data augmentations respectively.}
	\label{fig:learning_curves_all}
\end{figure*}

In Fig. \ref{fig:learning_curves_all}, we examine the learning process of fixed-dataset periodic training and training on non-repetitive samples based on ResNet-50 with no, weak, and strong data augmentations. To this end, we plot the evolving curves of the following eight quantities with the training: training loss measured on the synthetic training set, test loss (IID) measured on the synthetic IID test set, training accuracy measured on the synthetic training set, test accuracy (IID) measured on the synthetic IID test set, test loss (IID w/o BG) measured on the synthetic IID without background test set, test loss (OOD) measured on the SubVisDA-10 real/OOD test set, test accuracy (IID w/o BG) measured on the synthetic IID without background test set, and test accuracy (OOD) measured on the SubVisDA-10 real/OOD test set. The accuracy is measured using the ground truth labels, just for visualization.

\subsection{Visualizing Saliency Map}
\label{sec:saliency_maps}

\begin{figure*}[!ht]
	\centering
	\subfloat[Fixed]{
		\begin{minipage}[t]{0.5\textwidth}
			\centering
			\includegraphics[height=3.8in, width=2.3in]{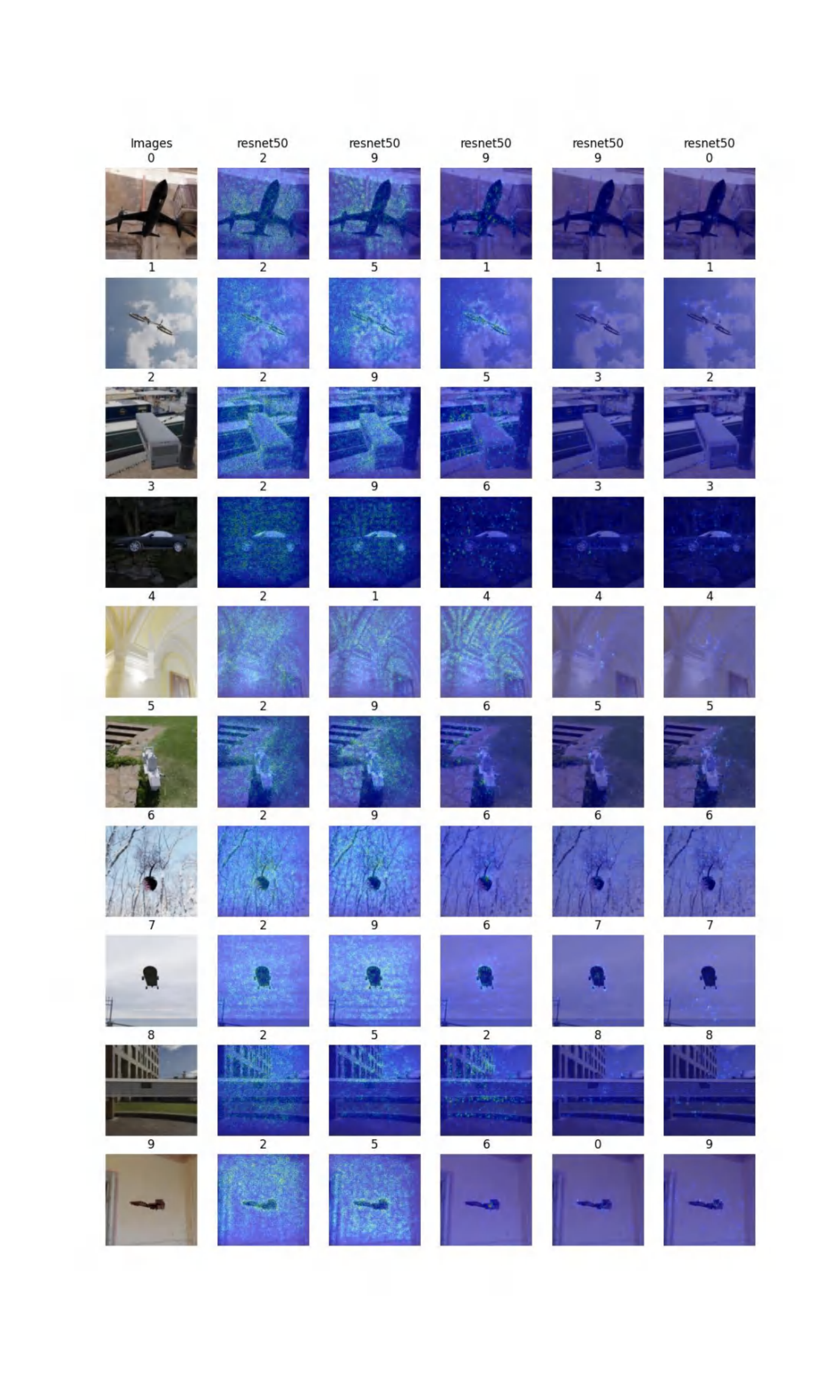}
			\label{fig:saliency_maps_resnet:1}
		\end{minipage}
	}
	\subfloat[Non-repetitive]{
		\begin{minipage}[t]{0.5\textwidth}
			\centering
			\includegraphics[height=3.8in, width=2.3in]{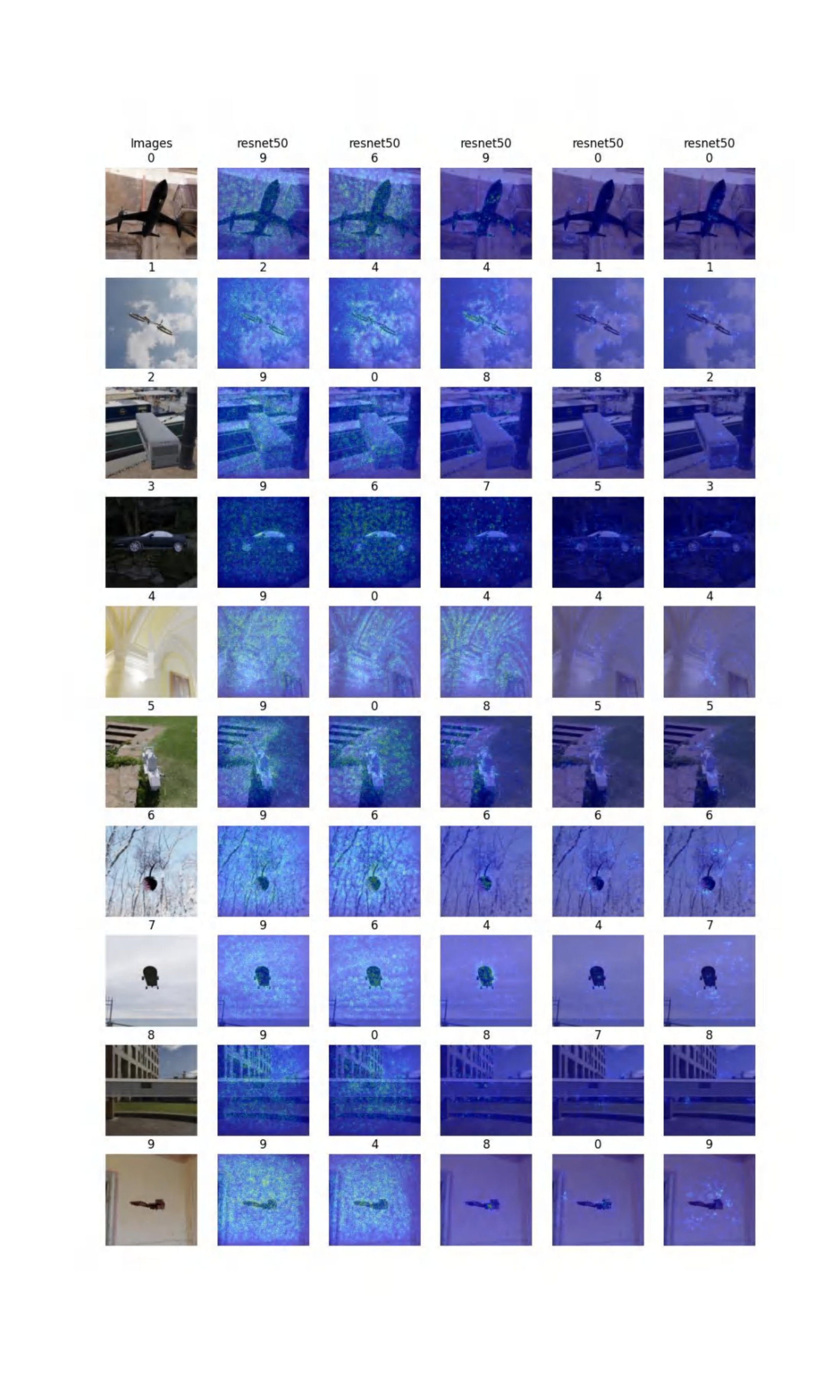}
			\label{fig:saliency_maps_resnet:2}
		\end{minipage}
	}
	\\
	\subfloat[Fixed]{
		\begin{minipage}[t]{0.5\textwidth}
			\centering
			\includegraphics[height=3.8in, width=2.3in]{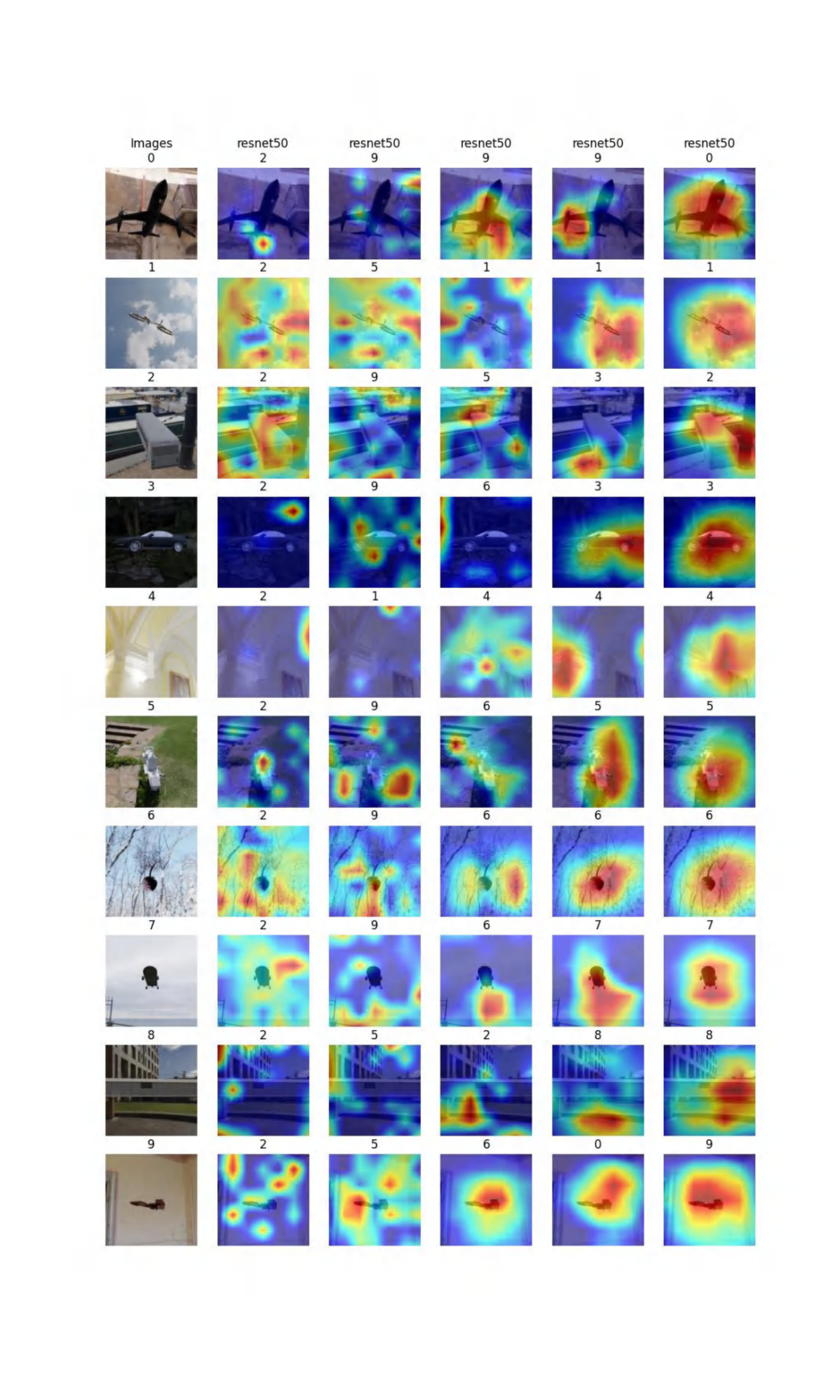}
			\label{fig:saliency_maps_resnet:3}
		\end{minipage}
	}
	\subfloat[Non-repetitive]{
		\begin{minipage}[t]{0.5\textwidth}
			\centering
			\includegraphics[height=3.8in, width=2.3in]{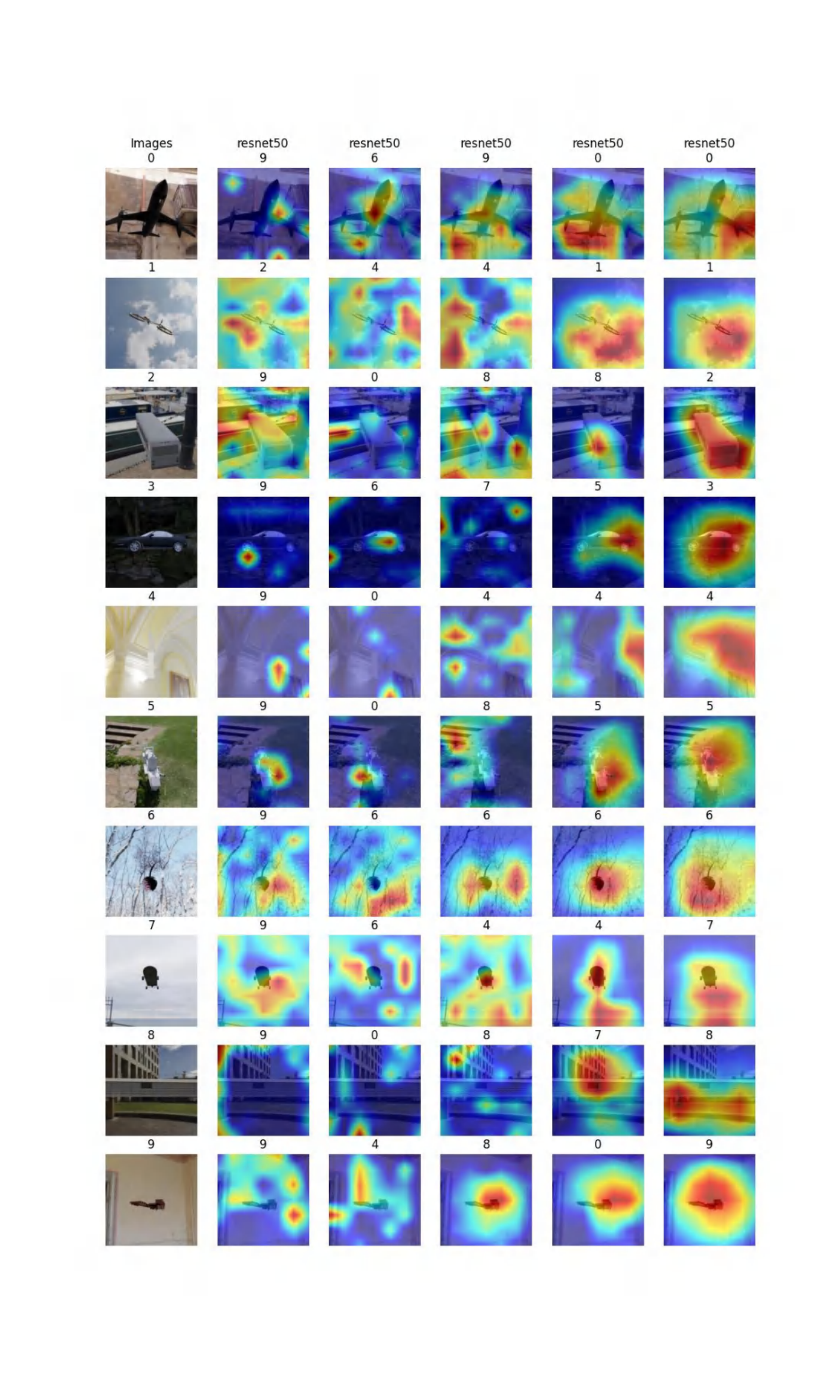}
			\label{fig:saliency_maps_resnet:4}
		\end{minipage}
	}
	\vspace{-0.3cm}
	\caption{Saliency maps of randomly selected IID test samples, obtained from the ResNet-50 trained on a fixed dataset or non-repetitive samples with no data augmentation, at the $20$-th, $200$-th, $2$K-th, $20$K-th, and $200$K-th training iterations. Note that rows 1 and 2 show input gradients \cite{input_grad} and gradient weighted class activation maps \cite{grad_cam} on input images respectively; the number on top of each picture means the ground-truth (first column) or predicted labels (other columns).}
	\label{fig:saliency_maps_resnet}
\end{figure*}

\begin{figure*}[!ht]
	\centering
	\subfloat[Fixed]{
		\begin{minipage}[t]{0.5\textwidth}
			\centering
			\includegraphics[height=3.8in, width=2.3in]{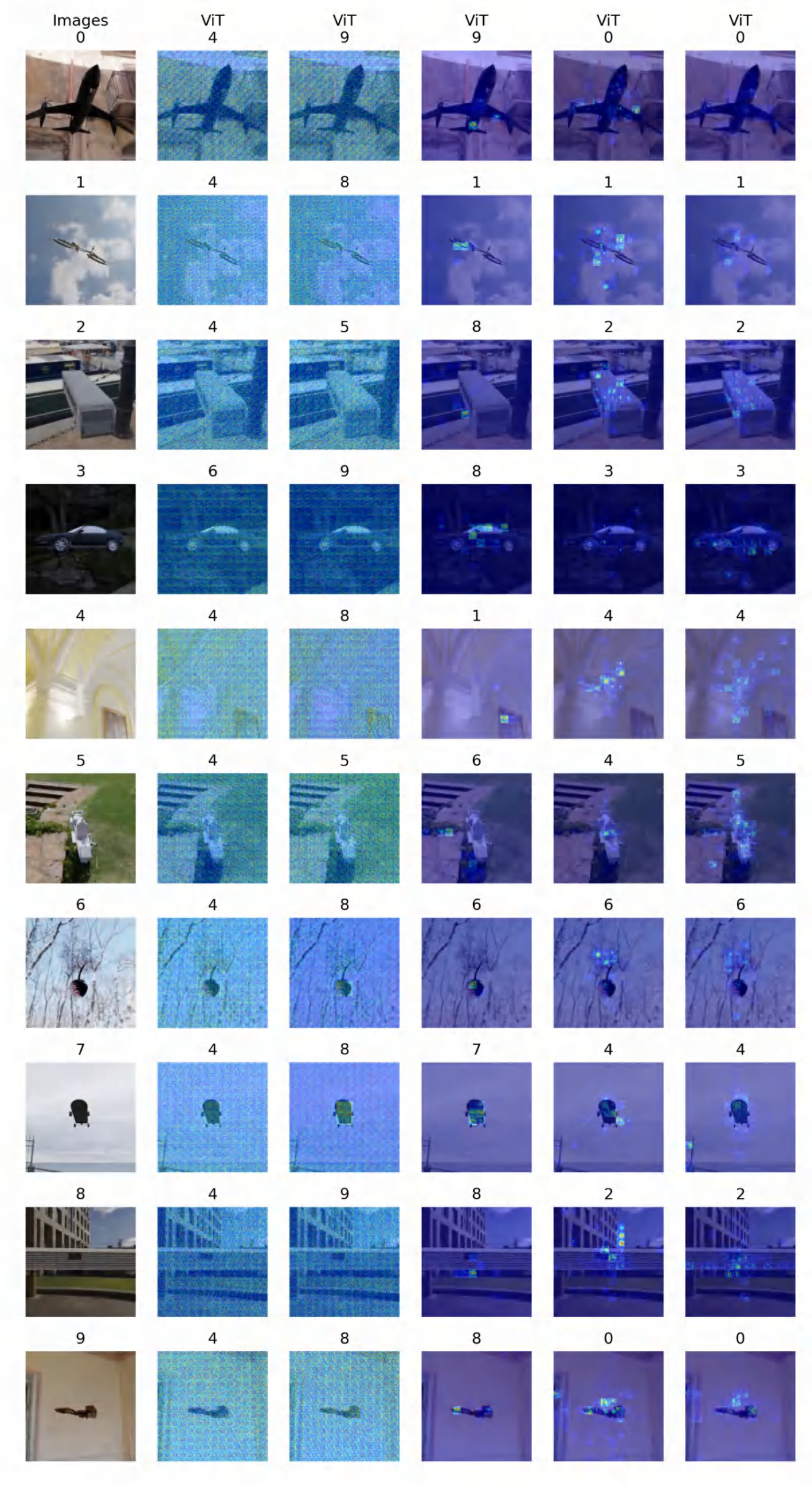}
			\label{fig:saliency_maps_vit:1}
		\end{minipage}
	}
	\subfloat[Non-repetitive]{
		\begin{minipage}[t]{0.5\textwidth}
			\centering
			\includegraphics[height=3.8in, width=2.3in]{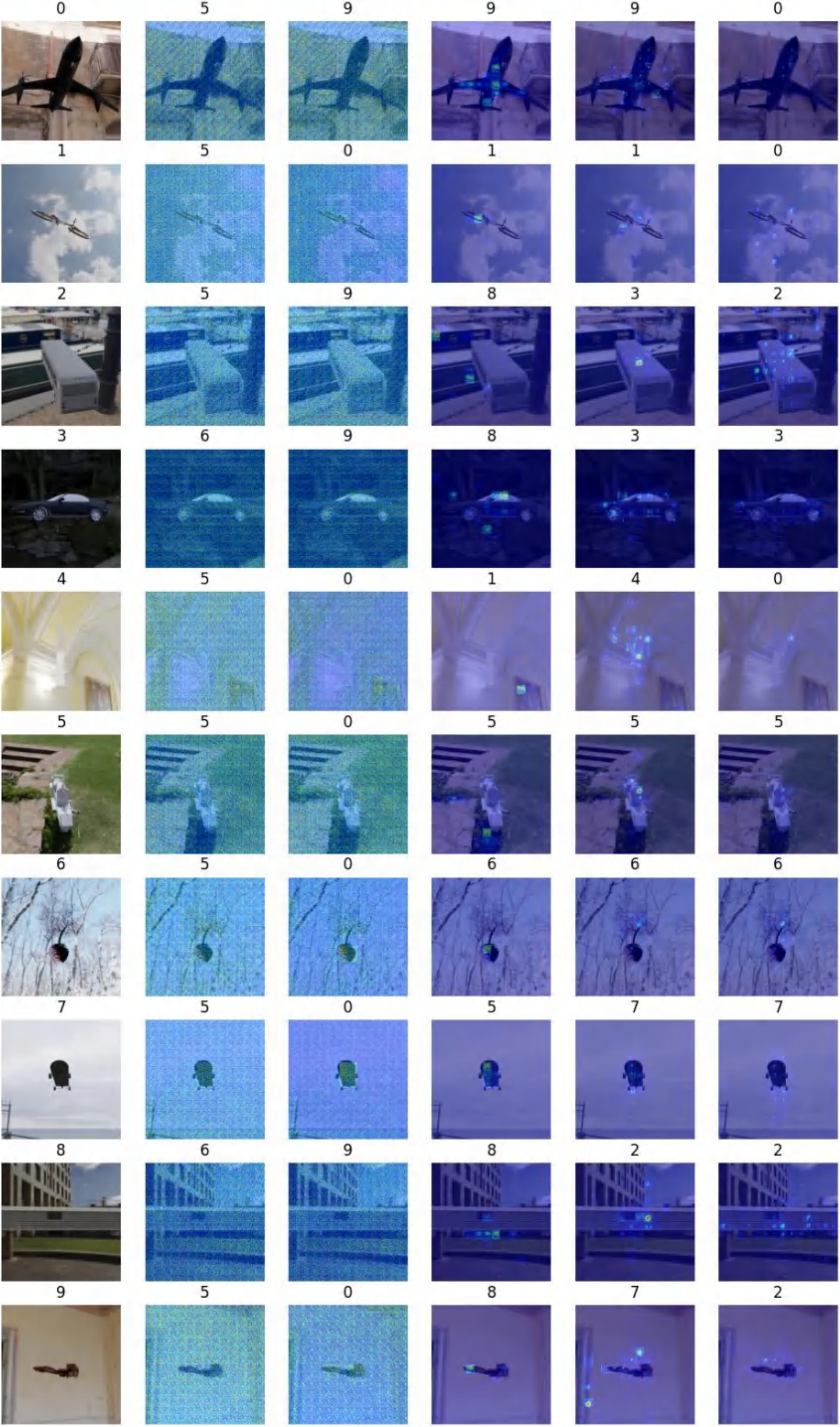}
			\label{fig:saliency_maps_vit:2}
		\end{minipage}
	}
	\\
	\subfloat[Fixed]{
		\begin{minipage}[t]{0.5\textwidth}
			\centering
			\includegraphics[height=3.8in, width=2.3in]{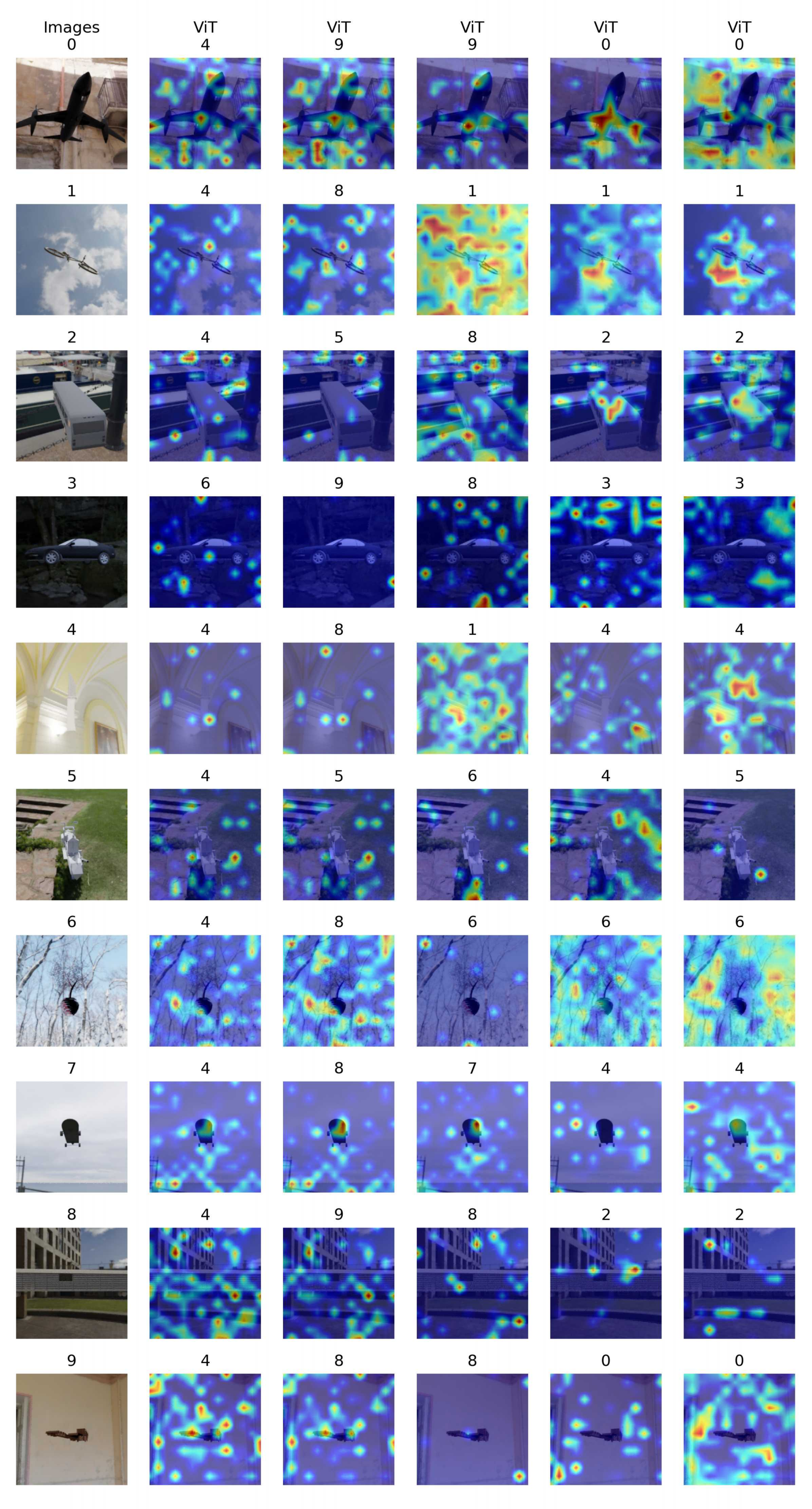}
			\label{fig:saliency_maps_vit:3}
		\end{minipage}
	}
	\subfloat[Non-repetitive]{
		\begin{minipage}[t]{0.5\textwidth}
			\centering
			\includegraphics[height=3.8in, width=2.3in]{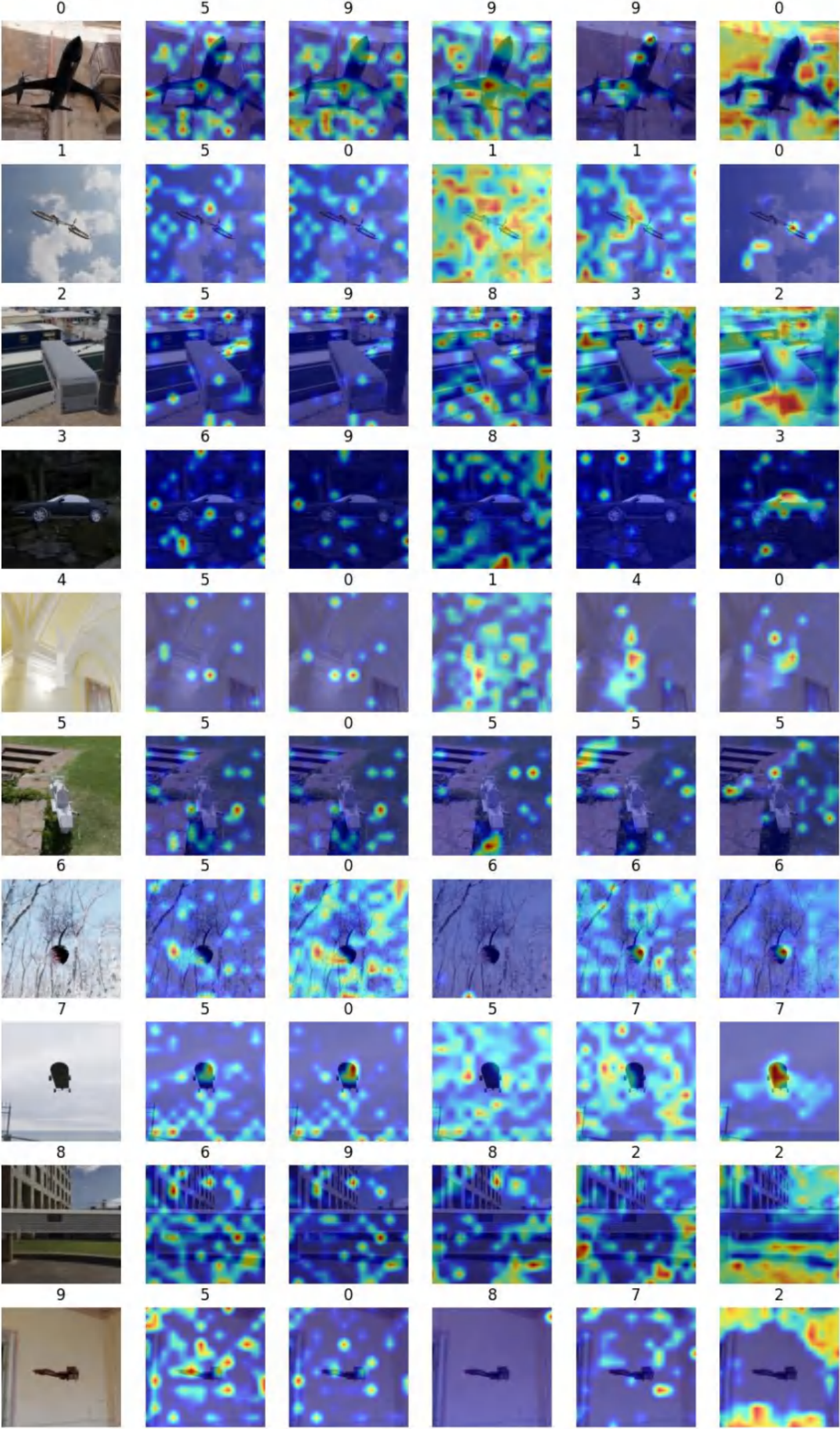}
			\label{fig:saliency_maps_vit:4}
		\end{minipage}
	}
	\vspace{-0.3cm}
	\caption{Saliency maps of randomly selected IID test samples, obtained from the ViT-B trained on a fixed dataset or non-repetitive samples with no data augmentation, at the $20$-th, $200$-th, $2$K-th, $20$K-th, and $200$K-th training iterations. Note that rows 1 and 2 show input gradients \cite{input_grad} and gradient weighted class activation maps \cite{grad_cam} on input images respectively; the number on top of each picture means the ground-truth (first column) or predicted labels (other columns).}
	\label{fig:saliency_maps_vit}
\end{figure*}

\begin{figure*}[!ht]
	\centering
	\subfloat[Fixed]{
		\begin{minipage}[t]{0.5\textwidth}
			\centering
			\includegraphics[height=3.8in, width=2.3in]{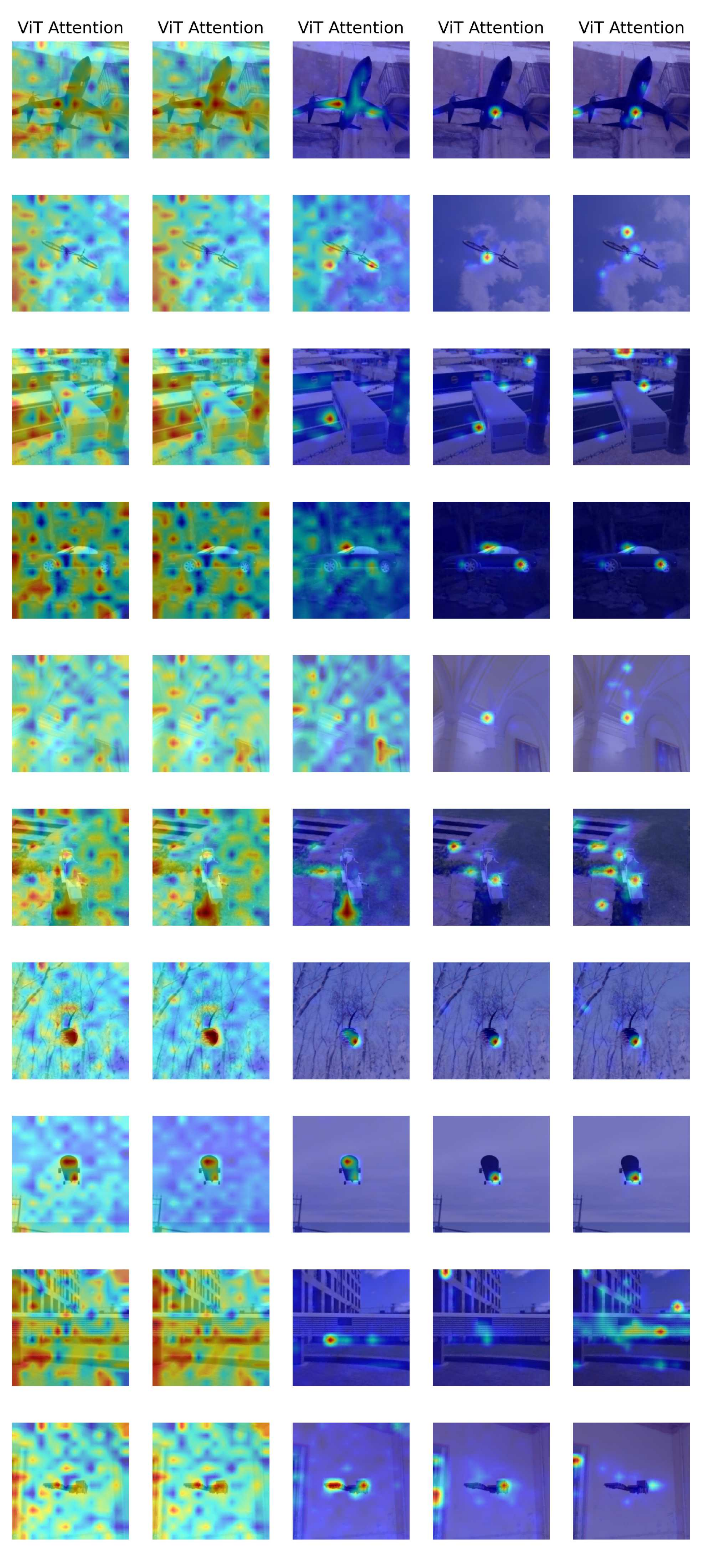}
			\label{fig:attn_maps_vit:1}
		\end{minipage}
	}
	\subfloat[Non-repetitive]{
		\begin{minipage}[t]{0.5\textwidth}
			\centering
			\includegraphics[height=3.8in, width=2.3in]{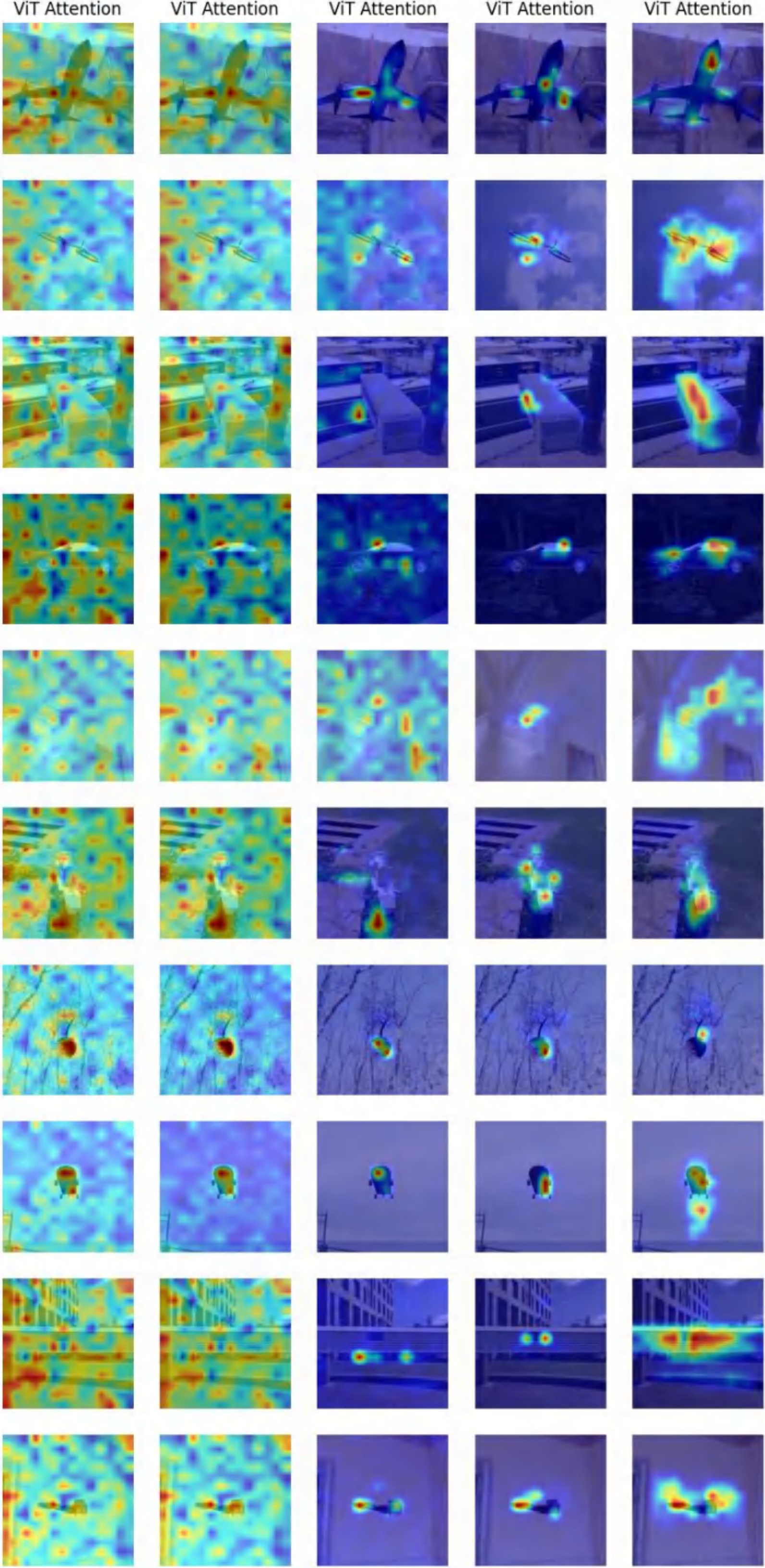}
			\label{fig:attn_maps_vit:2}
		\end{minipage}
	}
	\vspace{-0.3cm}
	\caption{Attention maps \cite{attn_flow} of randomly selected IID test samples, obtained from the ViT-B trained on a fixed dataset or non-repetitive samples with no data augmentation, at the $20$-th, $200$-th, $2$K-th, $20$K-th, and $200$K-th training iterations.}
	\label{fig:attn_maps_vit}
\end{figure*}

\begin{figure*}[!ht]
	\centering
	\subfloat[Fixed]{
		\begin{minipage}[t]{0.5\textwidth}
			\centering
			\includegraphics[height=3.8in, width=2.3in]{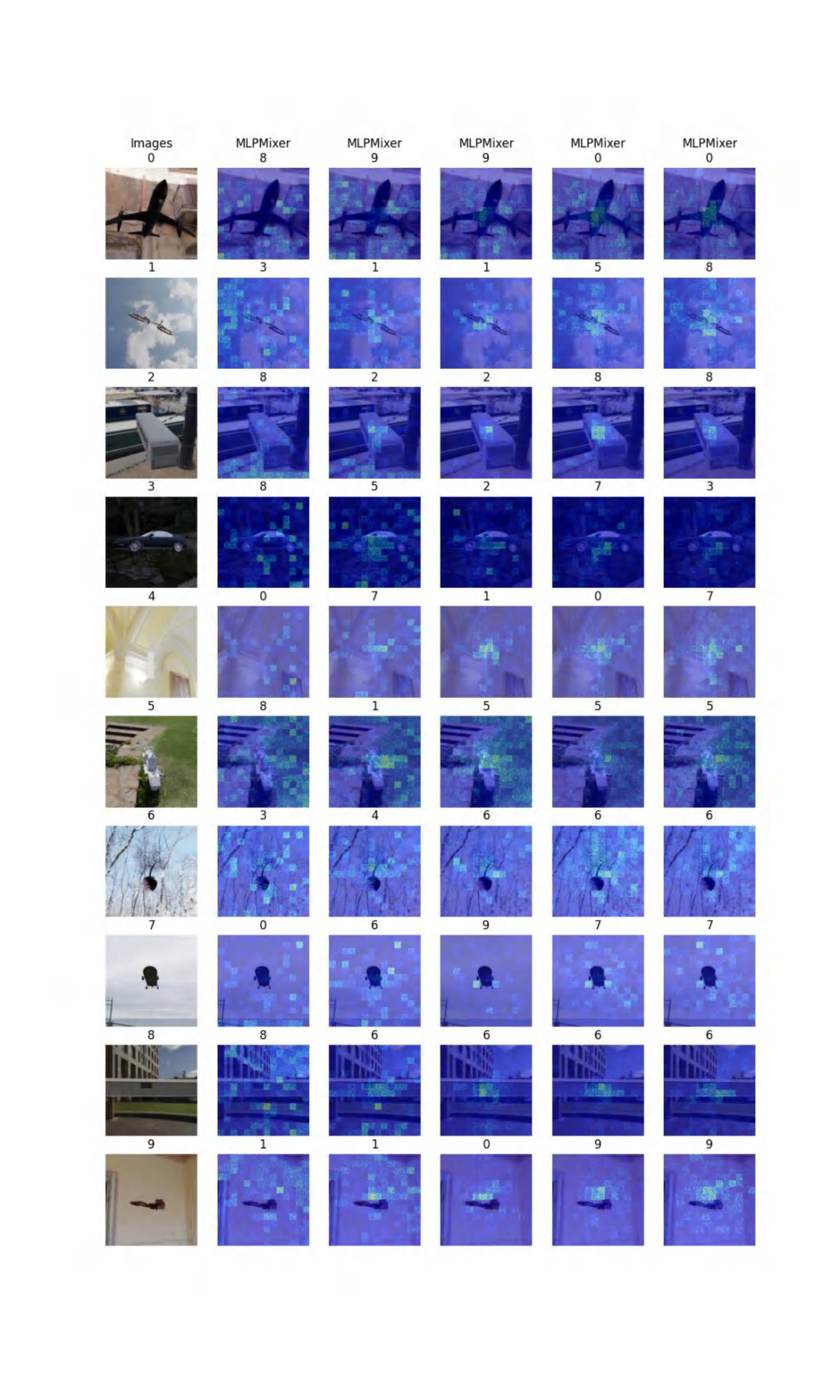}
			\label{fig:saliency_maps_mixer:1}
		\end{minipage}
	}
	\subfloat[Non-repetitive]{
		\begin{minipage}[t]{0.5\textwidth}
			\centering
			\includegraphics[height=3.8in, width=2.3in]{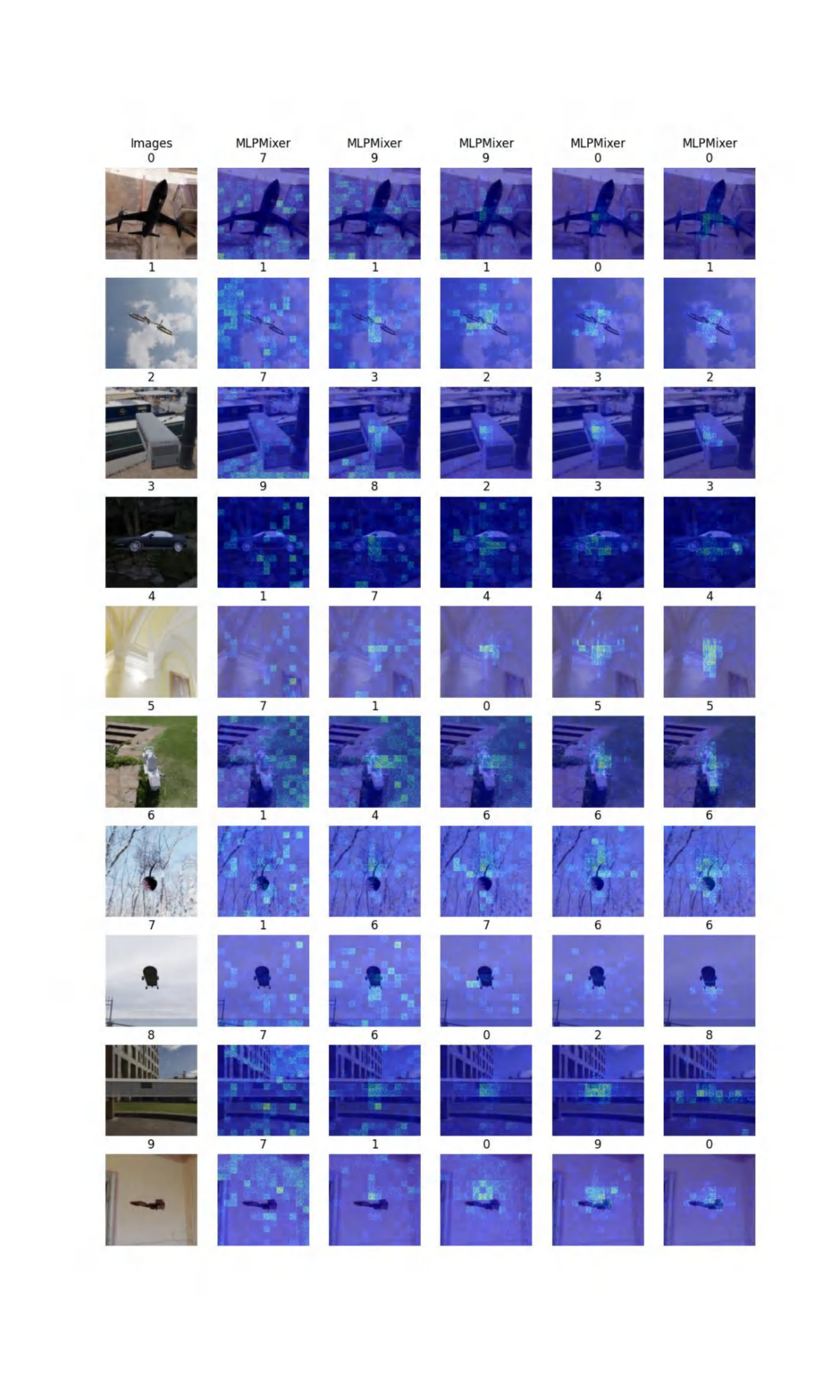}
			\label{fig:saliency_maps_mixer:2}
		\end{minipage}
	}
	\\
	\subfloat[Fixed]{
		\begin{minipage}[t]{0.5\textwidth}
			\centering
			\includegraphics[height=3.8in, width=2.3in]{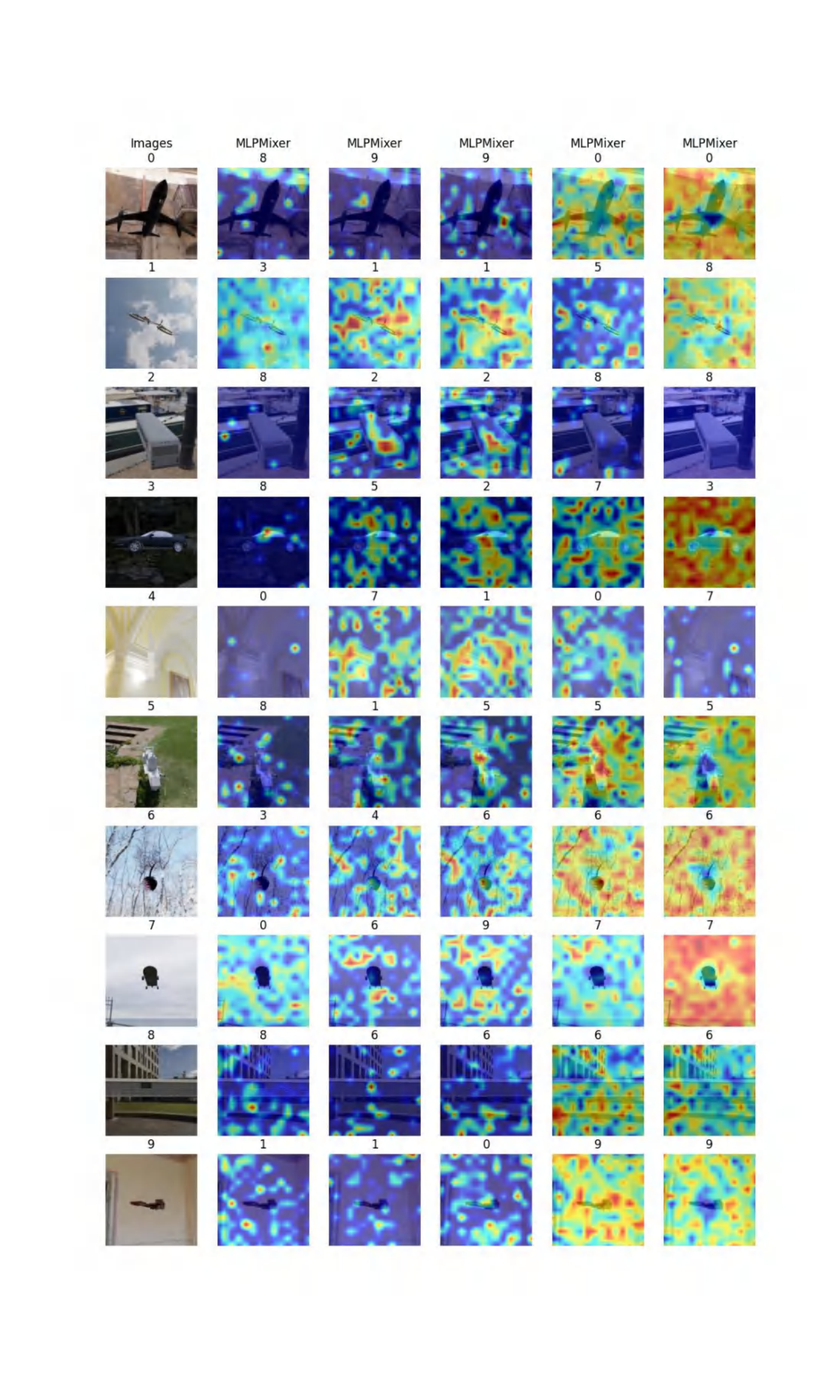}
			\label{fig:saliency_maps_mixer:3}
		\end{minipage}
	}
	\subfloat[Non-repetitive]{
		\begin{minipage}[t]{0.5\textwidth}
			\centering
			\includegraphics[height=3.8in, width=2.3in]{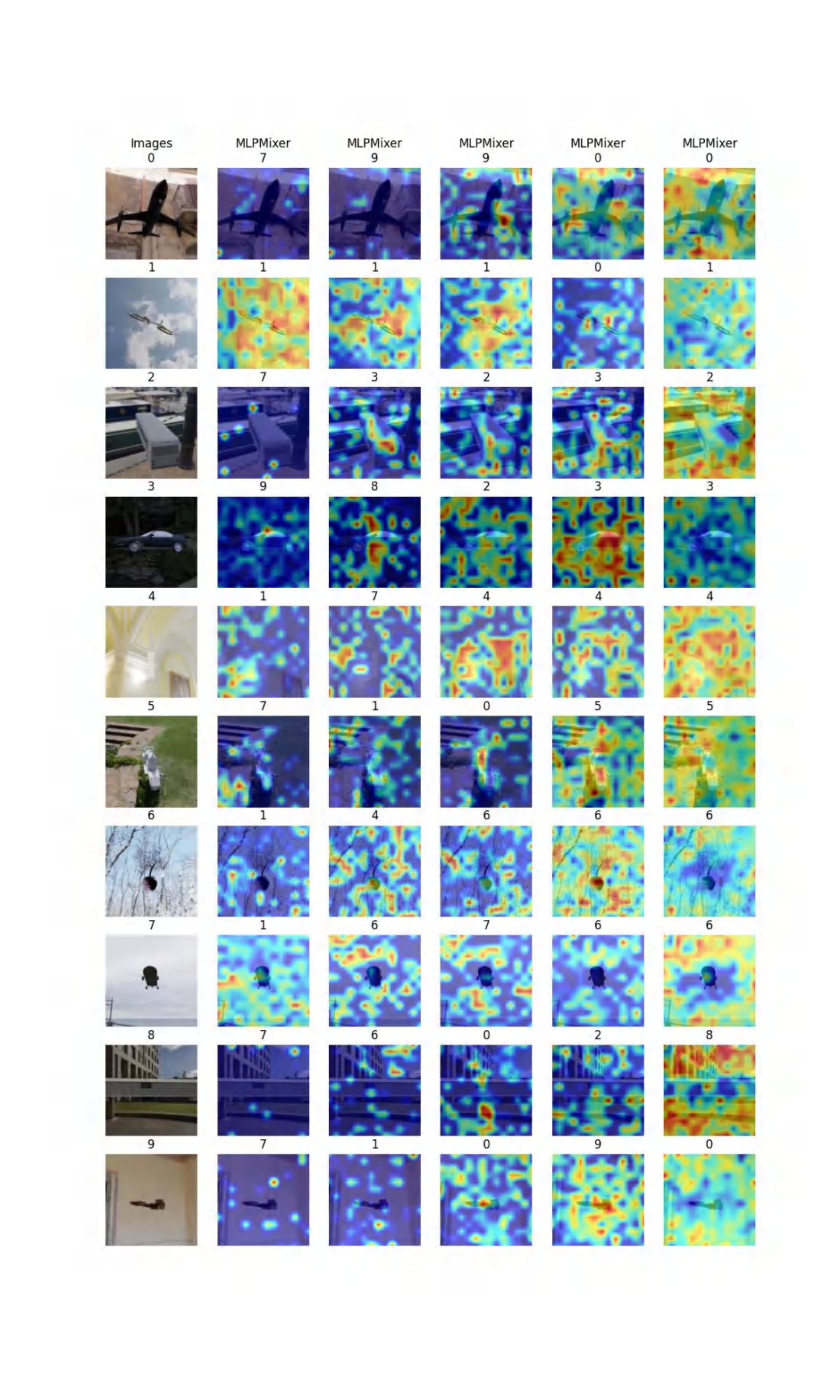}
			\label{fig:saliency_maps_mixer:4}
		\end{minipage}
	}
	\vspace{-0.3cm}
	\caption{Saliency maps of randomly selected IID test samples, obtained from the Mixer-B trained on a fixed dataset or non-repetitive samples with no data augmentation, at the $20$-th, $200$-th, $2$K-th, $20$K-th, and $200$K-th training iterations. Note that rows 1 and 2 show input gradients \cite{input_grad} and gradient weighted class activation maps \cite{grad_cam} on input images respectively; the number on top of each picture means the ground-truth (first column) or predicted labels (other columns).}
	\label{fig:saliency_maps_mixer}
\end{figure*}

We visualize the saliency maps, obtained from the ResNet-50 (Fig. \ref{fig:saliency_maps_resnet}), ViT-B (Fig. \ref{fig:saliency_maps_vit}), and Mixer-B (Fig. \ref{fig:saliency_maps_mixer}) trained on a fixed dataset or non-repetitive samples with no data augmentation. We consider two types of saliency visualization methods: input gradients which backpropagates the output score at the ground-truth category to the input image, and Grad-CAM which weights the feature maps with the gradients w.r.t. the features. For ViT-B, in Fig. \ref{fig:attn_maps_vit}, we also visualize the attention maps of the classification token to all image patches at the last multi-head self-attention layer. The last five columns correspond to results at the $20$-th, $200$-th, $2$K-th, $20$K-th, and $200$K-th training iterations respectively. The example image in each row is randomly selected from IID test data.

\subsection{More Impact of Data Augmentations}
\label{sec:impact_data_aug}

From Table 1, in OOD tests, training on non-repetitive images with no augmentation is superior to the fixed-dataset periodic training with weak augmentation, but far inferior to that with strong augmentation. It to some extent implies that the image transformations produced by 3D rendering itself do contain the hand-crafted weak augmentation changing pixel position in an image, but not the strong one changing both position and value of pixels.

\section{Evaluating Various Network Architectures}
\label{sec:compare_3_models}

\begin{figure*}[!ht]
	\centering
	\subfloat[Training loss]{
		\begin{minipage}[t]{0.25\textwidth}
			\includegraphics[height=1.05in]{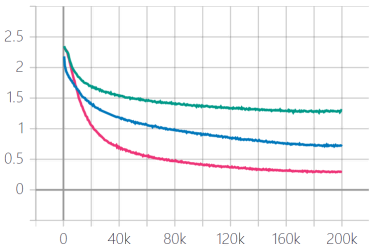}
			\label{fig:comparison_of_3models_all:1}
		\end{minipage}
	}
	\subfloat[Test loss (IID)]{
		\begin{minipage}[t]{0.25\textwidth}
			\includegraphics[height=1.05in]{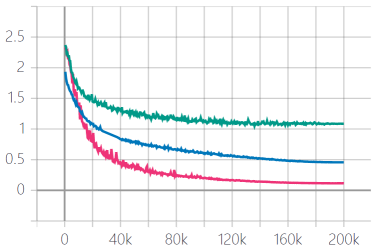}
			\label{fig:comparison_of_3models_all:2}
		\end{minipage}
	}
	\subfloat[Training acc.]{
		\begin{minipage}[t]{0.25\textwidth}
			\includegraphics[height=1.05in]{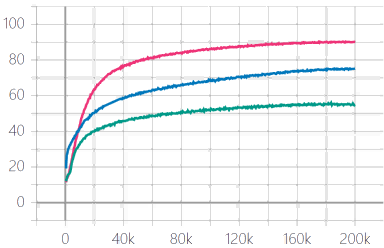}
			\label{fig:comparison_of_3models_all:3}
		\end{minipage}
	}
	\subfloat[Test acc. (IID)]{
		\begin{minipage}[t]{0.26\textwidth}
			\includegraphics[height=1.051in]{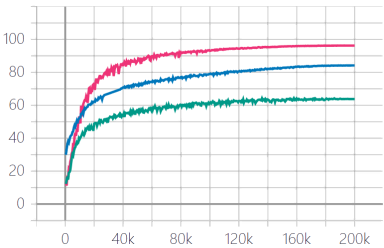}
			\label{fig:comparison_of_3models_all:4}
		\end{minipage}
	}
	\\
	\subfloat[Test loss (IID w/o BG)]{
		\begin{minipage}[t]{0.25\textwidth}
			\includegraphics[height=1.05in]{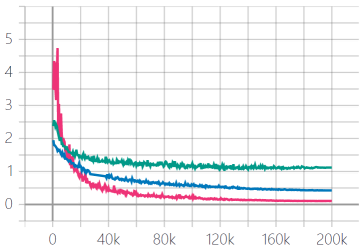}
			\label{fig:comparison_of_3models_all:5}
		\end{minipage}
	}
	\subfloat[Test loss (OOD)]{
		\begin{minipage}[t]{0.25\textwidth}
			\includegraphics[height=1.05in]{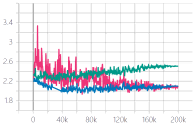}
			\label{fig:comparison_of_3models_all:6}
		\end{minipage}
	}
	\subfloat[Test acc. (IID w/o BG)]{
		\begin{minipage}[t]{0.25\textwidth}
			\includegraphics[height=1.05in]{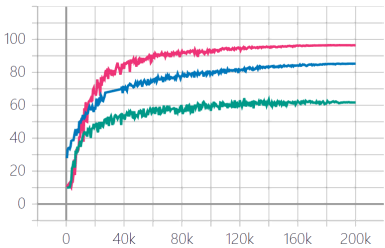}
			\label{fig:comparison_of_3models_all:7}
		\end{minipage}
	}
	\subfloat[Test acc. (OOD)]{
		\begin{minipage}[t]{0.26\textwidth}
			\includegraphics[height=1.05in]{images/comparison_of_3models/test_acc1_real_3models_sda.svg}
			\label{fig:comparison_of_3models_all:8}
		\end{minipage}
	}	
	\caption{Learning process of training various network architectures on non-repetitive samples with strong data augmentation. Note that \textcolor{red}{red}, \textcolor{green}{green}, and \textcolor{blue}{blue} indicate ResNet-50, ViT-B, and Mixer-B respectively.}
	\label{fig:comparison_of_3models_all}
\end{figure*}

In Fig. \ref{fig:comparison_of_3models_all}, we evaluate various network architectures by plotting their learning curves, in terms of training loss, test loss (IID), training accuracy, test accuracy (IID), test loss (IID w/o BG), test loss (OOD),  test accuracy (IID w/o BG), and test accuracy (OOD). Various network architectures --- ResNet-50 \cite{resnet}, ViT-B \cite{ViT}, and Mixer-B \cite{MLP-Mixer} are trained on non-repetitive samples with strong data augmentation.

We note that ViT performs poorly despite data augmentation and more training epochs. It may be because ViT cannot well fit datasets with high dimension and large variance, and does not learn features less dependent on the background. Two possible solutions to improve are decreasing the input patch size for fine-grained feature interaction and smoothing the dataset (e.g. mixup \cite{mixup}) for data distribution completeness.

\section{Comparing Pre-training for Domain Adaptation}
\label{sec:compare_pretrain_for_da}

\begin{figure*}[!ht]
	\centering
	\subfloat[No Adaptation]{
		\begin{minipage}[t]{0.33\textwidth}
			\centering
			\includegraphics[height=1.5in]{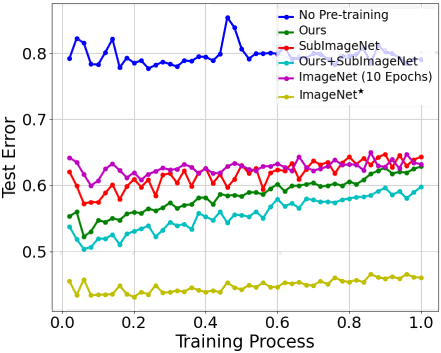}
			\label{fig:da_learning_curves:1}
		\end{minipage}
	}
	\subfloat[DANN]{
		\begin{minipage}[t]{0.33\textwidth}
			\centering
			\includegraphics[height=1.5in]{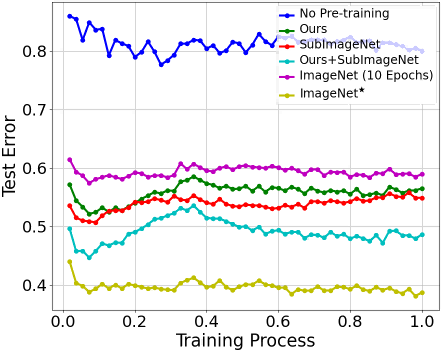}
			\label{fig:da_learning_curves:2}
		\end{minipage}
	}
	\subfloat[MCD]{
		\begin{minipage}[t]{0.33\textwidth}
			\centering
			\includegraphics[height=1.5in]{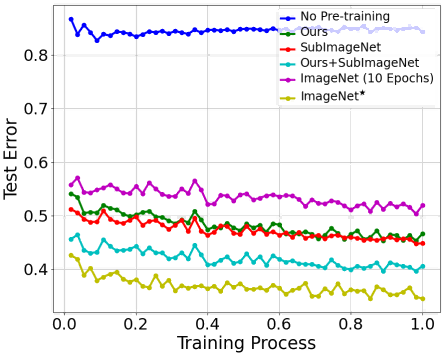}
			\label{fig:da_learning_curves_all:3}
		\end{minipage}
	}
	\\
	\subfloat[RCA]{
		\begin{minipage}[t]{0.33\textwidth}
			\centering
			\includegraphics[height=1.5in]{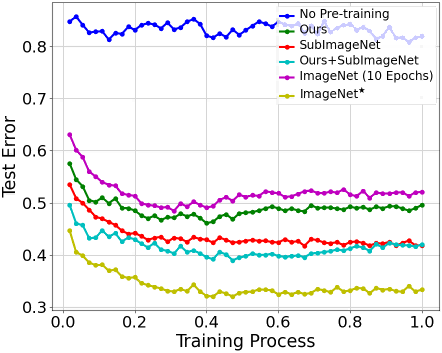}
			\label{fig:da_learning_curves:4}
		\end{minipage}
	}
	\subfloat[SRDC]{
		\begin{minipage}[t]{0.33\textwidth}
			\centering
			\includegraphics[height=1.5in]{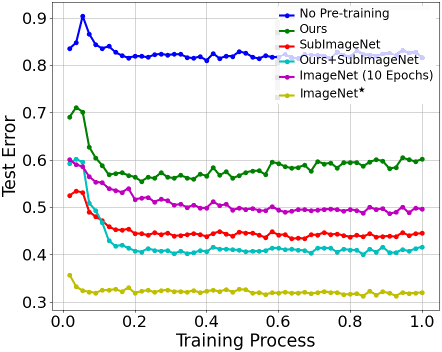}
			\label{fig:da_learning_curves:5}
		\end{minipage}
	}
	\subfloat[DisClusterDA]{
		\begin{minipage}[t]{0.33\textwidth}
			\centering
			\includegraphics[height=1.5in]{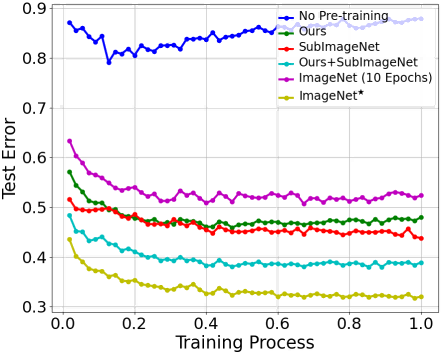}
			\label{fig:da_learning_curves_all:6}
		\end{minipage}
	}
	\caption{Learning process (Acc.) of domain adaptation when varying the pre-training scheme.}
	\label{fig:da_learning_curves_all}
\end{figure*}

In Fig. \ref{fig:da_learning_curves_all}, we compare different pre-training data using domain adaptation (DA) on SubVisDA-10 as the downstream task and show the learning process for several representative DA approaches. The considered pre-training schemes include \textbf{(1)} No Pre-training where the model parameters are randomly initialized, \textbf{(2)} Ours denotes our synthesized $120$K images of the $10$ object classes shared by SubVisDA-10, \textbf{(3)} SubImageNet is the subset collecting examples of the $10$ classes from ImageNet \cite{imagenet}, \textbf{(4)} ImageNet ($10$ Epoch) has 1K classes and $10$ pre-training epochs, and \textbf{(5)} ImageNet$^\bigstar$ uses the official ResNet-50 \cite{resnet} checkpoint pre-trained on ImageNet for $120$ epochs. The compared DA methods include No Adaptation that trains the model only on the labeled source data, DANN \cite{dann}, MCD, \cite{mcd}, RCA \cite{rca}, SRDC \cite{srdc}, and DisClusterDA \cite{disclusterda}.

\section{More Details on Our Proposed S2RDA Benchmark}
\label{sec:s2r}

\begin{figure*}[!ht]
	\centering
	\includegraphics[width=1.0\textwidth]{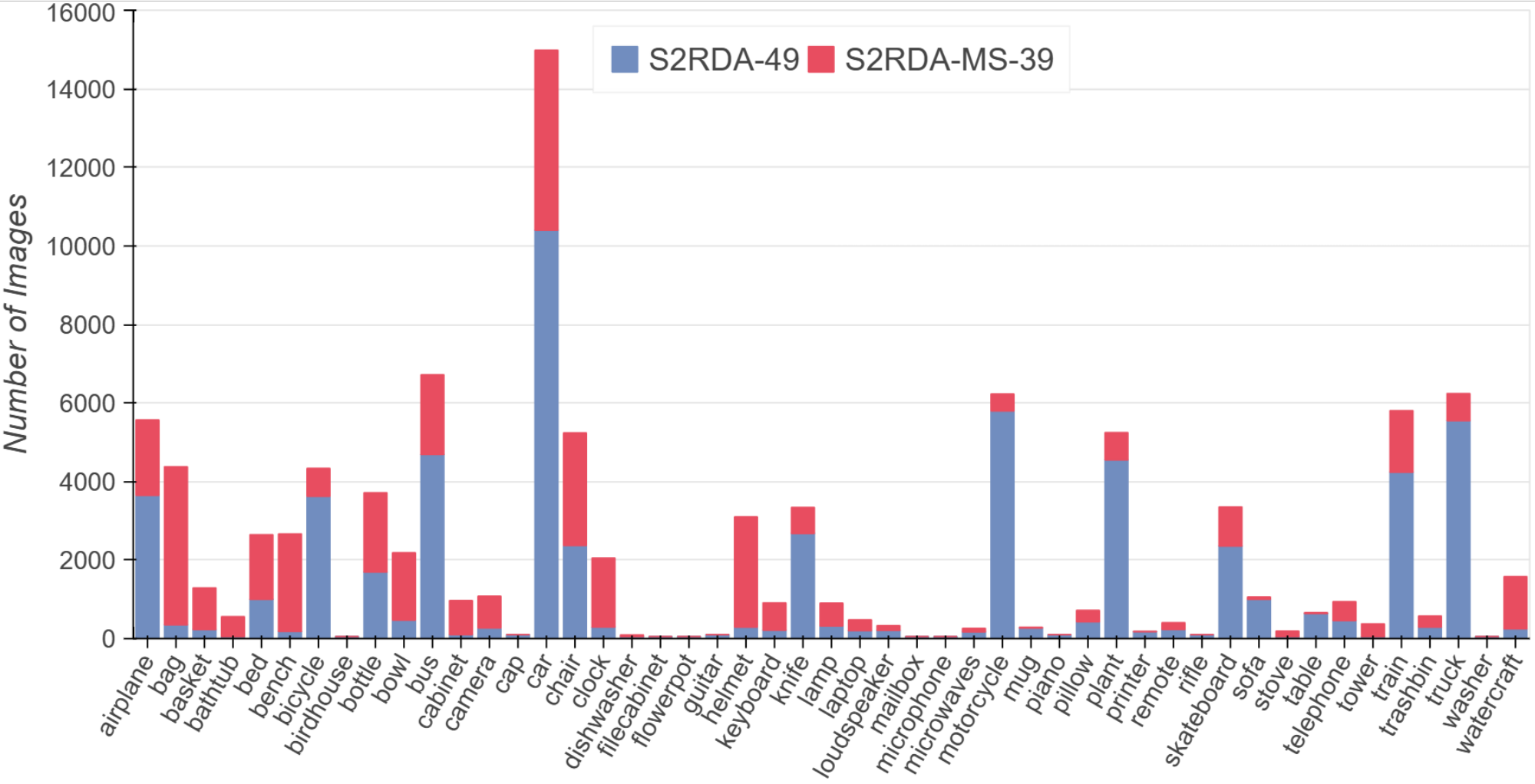}
	\caption{Distribution of the number of images per class in each real domain of our proposed S2RDA.}
	\label{fig:dataset_statistics}
\end{figure*}

\paragraph{Dataset Details.} 
Our proposed Synthetic-to-Real (S2RDA) benchmark for more practical visual domain adaptation (DA) includes two challenging transfer tasks of S2RDA-49 and S2RDA-MS-39. 
In Fig. \ref{fig:dataset_statistics}, we show the distribution of the number of images per class in each real domain, which is exhibited to be a long-tailed distribution where a small number of classes dominate. How we collect the real data from diverse real-world sources is recorded in respective files included in the code. Our S2RDA dataset is publicly available at \url{https://pan.baidu.com/s/1fHHaqrEHbUZLXEg9XKpgSg?pwd=w9wa}.

\paragraph{Comparing Synthetic Data with Real Data.} 
We provide quantitative and qualitative comparisons for VisDA-2017, our synthesized dataset, and ImageNet as follows. The mean and standard deviation of VisDA-2017, our synthesized dataset, and ImageNet are [0.878, 0.876, 0.874] and [0.207, 0.210, 0.216], [0.487, 0.450, 0.462] and [0.237, 0.251, 0.270], and [0.485, 0.456, 0.406] and [0.229, 0.224, 0.225] respectively. As we can see, the statistics of our synthesized dataset are closer to the real dataset ImageNet than VisDA-2017. It is consistent with the observation in Table 1 that our synthesized dataset used for training yields higher OOD/real test accuracy than SubVisDA-10. Except for quantitative comparisons, we have also provided the qualitative visualization for the three datasets and the proposed S2RDA benchmark in Fig. 1 and Fig. 6 respectively, which demonstrates that our synthesized dataset is visually more similar to ImageNet.


\section{Other Implementation Details}
\label{sec:other_details}

\paragraph{Supervised Learning/Pre-training.} For each backbone in Sec. 4.1, all its layers up to the second last one are used as the feature extractor and the neuron number of its last FC layer is set as $10$ to have the classifier. We use the cosine learning rate schedule: the learning rate is adjusted by $\eta_p = 0.5 \eta_0 (1 + \cos (\pi p))$, where $p$ is the process of training iterations normalized to be in $[0, 1]$ and the initial learning rate $\eta_0 = 0.01$. The momentum, weight decay, and random seed are set as $0.9$, $0.0001$, and $1$ respectively. In light of fairness, the final normalization operation uses the ImageNet statistics consistently for all experiments.

\paragraph{Downstream Domain Adaptation.} In domain adaptation training, we use all labeled source samples and all unlabeled target samples as the training data. In each base model, the last FC layer is replaced with a new task-specific FC layer as the classifier. We fine-tune the pre-trained layers and train the new layer from scratch, where the learning rate of the latter is $10$ times that of the former. The learning rate is adjusted by $\eta_p = \eta_0(1+\alpha p)^{-\beta}$, where $p$ denotes the training process of training epochs normalized to be in $[0, 1]$, the initial learning rate $\eta_0$ is $0.001$ for MCD and $0.0001$ for other methods, $\alpha = 10$, and $\beta = 0.75$. The momentum, weight decay, and random seed are set as $0.9$, $0.0001$, and $0$ respectively. By convention, strong and weak data augmentations are applied in pre-training and domain adaptation respectively. For domain adaptation on our proposed S2RDA benchmark, we use ResNet-50 as the backbone, which is initialized by the official ImageNet pre-trained checkpoint \cite{resnet}. The initial learning rate is set as $0.0001$ across all experiments. Other implementation details are the same as those described above. 

More details are as follows.
\begin{enumerate}
	\item For 3D rendering with domain randomization, the monocular camera default in BlenderProc \cite{blenderproc} is used in the 3D renderer. 
	
	\item SubVisDA-10 includes the following $10$ classes: airplane, bicycle, bus, car, knife, motorbike, plant, skateboard, train, and truck.
	
	\item MetaShift \cite{MetaShift} we use is a filtered version of $2559865$ images from $376$ classes by running the officially provided code of dataset construction. It is formed by setting a threshold for subset size ($>=25$) and subset number in one class ($>5$).
	
	\item Mean class precision is the average over recognition precisions of all classes. It is an indicator of class imbalance that different categories have different prediction accuracy. When it deviates from the overall accuracy in a test, class imbalance happens.
	
	\item In Table 3, the number highlighted by the green color indicates the best Acc. in each row (among all compared DA methods), the number underlined by the red color indicates the best Mean in each row (among all compared DA methods), and the bold number in each column indicates the best result among all considered pre-training schemes. In Table 4, the bold number highlighted by the green color indicates the best Acc. in each row (among all compared DA methods), and the bold number underlined by the red color indicates the best Mean in each row (among all compared DA methods).
	
	\item PyTorch \cite{pytorch} is used for implementation. Grid Search is used for hyperparameter tuning. We use an 8-GPU NVIDIA GeForce GTX 1080 and an 8-GPU NVIDIA Tesla M40 to run experiments. For the used assets, we have cited the corresponding references in the main paper, and we mention their licenses here: CCTextures under CC0 License, Haven under CC0 License, ShapeNet under a custom license, VisDA-2017 under MIT License, ImageNet under a custom license, and MetaShift under MIT License.
\end{enumerate}

\section{Other Related Works}
\label{sec:related_works}

\noindent\textbf{Real Datasets.} A lot of large-scale real datasets \cite{imagenet,imagenet-21k,JFT-300M,MetaShift,StylizedImageNet,ms-coco,L-Bird,WebFG} have harnessed and organized the explosive image data from Internet or the real world for deep learning of meaningful visual representations. For example, ImageNet \cite{imagenet} is a large-scale database of images built upon the backbone of the WordNet structure; ImageNet-1K, consisting of $1.28$M images from $1$K common object categories, which serves as the primary dataset for pre-training deep models for computer vision tasks. Barbu et al. \cite{ObjectNet} collect a large real-world test set for more realistic object recognition, ObjectNet, which has $50$K images and is bias-controlled. Ridnik et al. \cite{imagenet-21k} dedicatedly preprocess the full set of ImageNet --- ImageNet-21K with the WordNet hierarchical structure utilized, such that high-quality efficient pre-training on the resulted ImageNet-21K-P (of $12$M images) can be made for practical use. In \cite{JFT-300M}, JFT-300M of more than $375$M noisy labels for $300$M images is exploited to study the effects of pre-training on current vision tasks. Geirhos et al. \cite{StylizedImageNet} construct StylizedImageNet by replacing the object texture in an image with a random painting style via style transfer, to learn a shape-based representation. MS COCO \cite{ms-coco} has $328$K images, where considerably more object instances exist as compared to ImageNet, enabling deep models to learn precise 2D localization. 
MetaShift \cite{MetaShift} of $2.56$M natural images ($\sim400$ classes) is formed by context guided clustering of the images from GQA \cite{CleanedVG}, a cleaned version of Visual Genome \cite{VisualGenome} connecting language and vision. CCT-20 \cite{AnimalDataset} and PACS \cite{DGDataset} are designed to measure recognition generalization to novel visual domains. Some works \cite{L-Bird,WebFG} leverage free, numerous web data to benchmark or assist fine-grained recognition. Some small datasets are used as benchmarks for semi-supervised learning  \cite{cifar,svhn,stl10} or domain generalization \cite{AnimalDataset,DGDataset}.

\paragraph{Data Manipulation.} Deep models are hungry for more training data in that the generalization ability often relies on the quantity and diversity of training samples. To improve model generalization with a limited set of training data available, the cheapest and simplest way is data manipulation, which increases the sample diversity from two different perspectives of data augmentation and data generation. The former applies a series of random image transformations \cite{survey_data_aug} or appends adversarial examples at each iteration \cite{adv_data_aug}; the latter uses generative models to generate diverse and rich data such as Variational Auto-Encoder (VAE) \cite{M-ADA} and Generative Adversarial Network (GAN) \cite{ClusterGAN} or renders 3D object models into RGB images via domain randomization \cite{domain_randomization,SDR4CarDetection,DR4ObjectDetection}.

\paragraph{Deep Models.} Deep model has strong representational capacity in that they can learn powerful, hierarchical representations when trained on large amounts of data; they can be highly scalable from various aspects of architectural innovation, such as spatial exploitation, depth, multi-path, width, feature-map exploitation, channel boosting, and attention. Deep Convolutional Neural Networks (CNNs) have been popularized for decades in a wide range of computer vision tasks. A comprehensive survey for CNNs can be found in \cite{survey_deep_cnn}. The most commonly used CNN-based network architecture is ResNet \cite{resnet}, which reformulates the layers as learning residual functions concerning the layer inputs, instead of learning unreferenced functions. Recently, some new types of network architectures have emerged, such as ViT \cite{ViT} and MLP-Mixer \cite{MLP-Mixer}. 
ViT stacks a certain number of multi-head self-attention layers and is applied directly to sequences of fixed-size image patches. 
MLP-Mixer is based exclusively on multi-layer perceptrons (MLPs) and contains two types of MLP layers: one for mixing channels in individual image patches and one for mixing features across patches of different spatial locations. In this work, we experiment on the three representative types of networks.

\paragraph{Transfer Learning.} There has been a huge literature in the field of transfer learning \cite{survey_tl_early,survey_tl_recent,survey_transferability_in_dl}, where the paradigm of pre-training and then fine-tuning has made outstanding achievements in many deep learning applications. Extensive studies have been done for supervised pre-training \cite{how_transferable,better_imagenet_better_transfer,BiT,rethink_sup_pretrain,MLP_projector_for_pretrain,study_pretrain_for_DA}. For example, Yosinski et al. \cite{how_transferable} examine the transferability of features at different layers along the network; the relationship between ImageNet accuracy and transferability is evaluated in \cite{better_imagenet_better_transfer}; BiT \cite{BiT} provides a recipe of the minimal number of existing tricks for pre-training and downstream transferring; in \cite{MLP_projector_for_pretrain}, an MLP projector is added before the classifier to improve the transferability; LOOK \cite{rethink_sup_pretrain} solves the problem of overfitting upstream tasks by only allowing nearest neighbors to share the class label, in order to preserve the intra-class semantic difference; particularly, Kin et al. \cite{study_pretrain_for_DA} preliminarily study the effects of pre-training on domain transfer tasks, from the aspects of network architectures, size, pre-training loss, and datasets. Another popular branch of self-supervised learning is increasingly important for transfer learning. Previous works have proposed various pretext tasks, such as image inpainting and jigsaw puzzle \cite{survey_selfsup_pretrain}. Recent works concentrate on self-supervised/unsupervised pre-training \cite{survey_selfsup_pretrain,CDS,improve_unsup_pretrain,EMAN,SwAV,SimCLR,BYOL,MoCo} and have shown powerful transferability on multiple downstream tasks, comparable to supervised pre-training. They often rely on contrastive learning to learn visual representations of rich intra-class diversity \cite{mcr2}, e.g., contrasting feature embeddings \cite{CDS,EMAN,SimCLR,BYOL,MoCo} or cluster assignments \cite{SwAV} of anchor, positive, and negative instances. Note that CDS \cite{CDS} proposes a second self-supervised pre-training stage using the unlabeled downstream data from multiple domains, which applies instance discrimination not only in individual domains but also across domains. Also, many researchers are devoted to improving fine-tuning by leveraging the pre-trained ImageNet knowledge \cite{CSG}, using pre-training data for fine-tuning \cite{improve_ft_use_pretrain_data}, improving regularization and robustness \cite{improve_reg_rob}, adapting unfamiliar inputs \cite{input_tuning}, applying the easy two-step strategy of linear probing and then full fine-tuning \cite{bad_finetuning}, to name a few. 
Different from \cite{study_pretrain_for_DA}, we focus on the utility of synthetic data and take the first step towards clearing the cloud of mystery surrounding how different pre-training schemes including synthetic data pre-training affect the practical, large-scale synthetic-to-real adaptation.

\noindent\textbf{Domain Adaptation.} Domain adaptation is a developing field with a huge diversity of approaches. A popular strategy is to explicitly model and minimize the distribution shift between the source and target domains  \cite{dann,mcd,rca,SymNets,tpn,vicatda}, such that the domain-invariant features can be learned and thus the task classifier trained on the labeled source data can well generalize to the unlabeled target domain. 
DANN \cite{dann} aligns the source and target domains as a whole by domain-adversarial training, i.e., reversing the signal from a domain discriminator, but does not utilize the discriminative information from the target domain. 
MCD \cite{mcd} minimizes the maximum prediction discrepancy between two task classifiers to learn domain-invariant and class-discriminative features. 
RCA \cite{rca} implements the domain-adversarial training based on a joint domain-category classifier to learn class-level aligned features, i.e., invariant at corresponding classes of the two domains. 
Differently, works of another emerging strategy \cite{srdc,disclusterda,tat,bnm} take steps towards implicit domain adaptation, without explicit feature alignment that could hurt the intrinsic discriminative structures of target data. 
SRDC \cite{srdc} uncovers the intrinsic target discrimination via deep discriminative target clustering in both the output and feature spaces with structural source regularization hinging on the assumption of structural similarity across domains. 
DisClusterDA \cite{disclusterda} proposes a new clustering objective for discriminative clustering of target data with distilled informative source knowledge, based on a robust variant of entropy minimization, a soft Fisher-like criterion, and the cluster ordering via centroid classification. 
In this work, we consider these representative DA methods for the empirical study, and broader introductions to the rich literature are provided in \cite{survey_da,U-WILDS}.

\paragraph{OOD Generalization.} Out-of-distribution (OOD) generalization, i.e., domain generalization, assumes the access to single or multiple different but related domains and aims to generalize the learned model to an unseen test domain. A detailed review for recent advances in domain generalization is presented in \cite{survey_dg}, which categorizes the popular algorithms into three classes: data manipulation \cite{adv_data_aug,M-ADA}, representation learning \cite{PDEN,ASR-Norm}, and learning strategy \cite{MetaReg,M-ADA}. For example, adversarial examples are generated to learn robust models in \cite{adv_data_aug}; a progressive domain expansion subnetwork and a domain-invariant representation learning subnetwork are jointly learned to mutually benefit from each other in \cite{PDEN}; Balaji et al. \cite{MetaReg} adopt the meta-learning strategy to learn a regularizer that the model trained on one domain can well generalize to another domain.

\end{appendices}

\end{document}